\documentclass[journal]{IEEEtran}
\usepackage{natbib}
\setcitestyle{numbers,square}
\usepackage{graphicx}
\usepackage{algorithm}
\usepackage{amsmath}
\usepackage{booktabs}
\usepackage{multirow}
\usepackage{pifont}
\usepackage{siunitx}
\usepackage{makecell}
\usepackage{subcaption}
\usepackage{tabularx}
\usepackage{amssymb}
\usepackage{float} 
\usepackage{algpseudocode}

\ifCLASSINFOpdf
\else
\fi
\begin{document}
%
\title{EFSI-DETR: Efficient Frequency-Semantic Integration for Real-Time Small Object Detection in UAV Imagery}
%
%
%

\author{Yu Xia,
        Chang Liu, 
        Tianqi, Xiang,
        Yang Cong,~\IEEEmembership{Senior Member,~IEEE},
        Zhigang Tu,~\IEEEmembership{Senior Member,~IEEE}
        
\thanks{Yu Xia, Tianqi Xiang and Zhigang Tu are with the State Key Laboratory of Information Engineering in Surveying, Mapping and Remote Sensing, Wuhan University, Wuhan, China. Zhigang Tu is also with the Wuhan University Shenzhen Research Institute, Shenzhen, China.}

\thanks{Chang Liu is with School of Computer Science, Wuhan University, China.}

\thanks{Yang Cong is with the School of Automation Science and Engineering, South China University of Technology, Guangzhou, China.}

\thanks{Yu Xia and Chang Liu contributed equally.}
\thanks{Corresponding author: Zhigang Tu (Email: tuzhigang@whu.edu.cn).}

}

%
%

\markboth{Journal of \LaTeX\ Class Files,~Vol.~14, No.~8, August~2025}%
{Shell \MakeLowercase{\textit{et al.}}: EFSI-DETR: Efficient Frequency-Semantic Integration for Real-Time Small Object Detection in UAV Imagery}
%



\maketitle

\begin{abstract}
Real-time small object detection in Unmanned Aerial Vehicle (UAV) imagery remains challenging due to limited feature representation and ineffective multi-scale fusion. Existing methods underutilize frequency information and rely on static convolutional operations, which constrain the capacity to obtain rich feature representations and hinder the effective exploitation of deep semantic features. To address these issues, we propose EFSI-DETR, a novel detection framework that integrates efficient semantic feature enhancement with dynamic frequency-spatial guidance. EFSI-DETR comprises two main components: (1) a Dynamic Frequency-Spatial Unified Synergy Network (DyFusNet) that jointly exploits frequency and spatial cues for robust multi-scale feature fusion, (2) an Efficient Semantic Feature Concentrator (ESFC) that enables deep semantic extraction with minimal computational cost. Furthermore, a Fine-grained Feature Retention (FFR) strategy is adopted to incorporate spatially rich shallow features during fusion to preserve fine-grained details, crucial for small object detection in UAV imagery. Extensive experiments on VisDrone and CODrone benchmarks demonstrate that our EFSI-DETR achieves the state-of-the-art performance with real-time efficiency, yielding improvement of \textbf{1.6}\% and \textbf{5.8}\% in AP and AP$_{s}$ on VisDrone, while obtaining \textbf{188} FPS inference speed on a single RTX 4090 GPU.
\end{abstract}


%
\IEEEpeerreviewmaketitle

\section{Introduction}
\begin{figure}[thb] \centering
      \begin{subfigure}[t]{0.24\textwidth}
        \includegraphics[width=\textwidth]{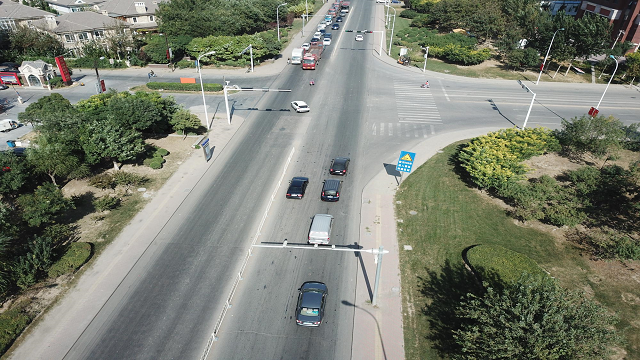}
        \includegraphics[width=\textwidth]{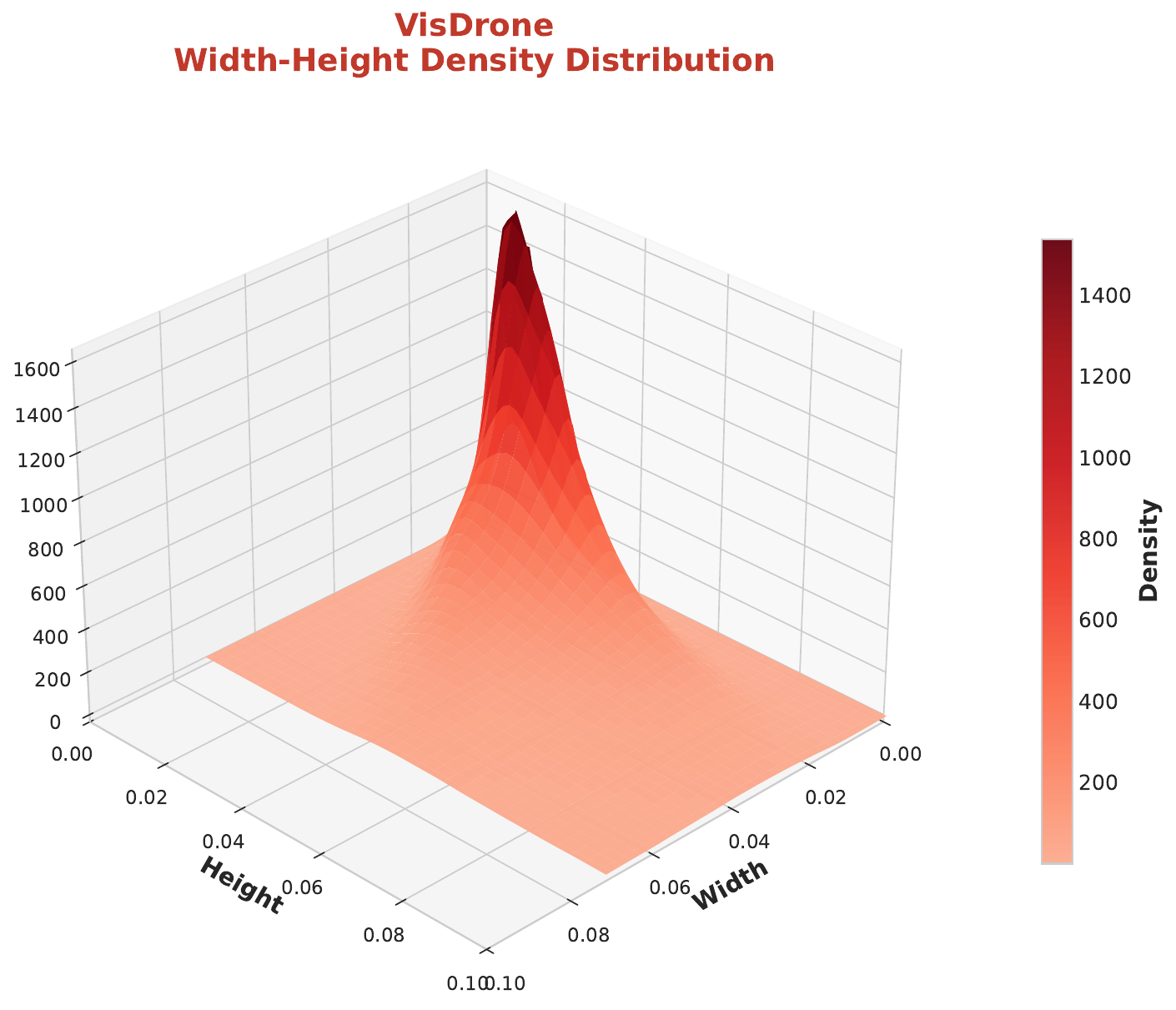}
        \caption*{\textbf{VisDrone}}\vspace{-1mm}
      \end{subfigure}\hspace{-1mm}
      \begin{subfigure}[t]{0.24\textwidth}
        \includegraphics[width=\textwidth]{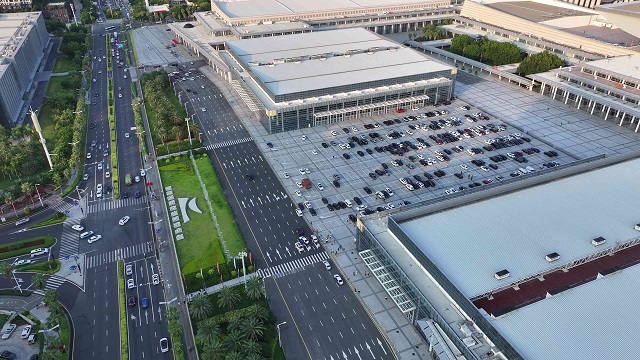}
        \includegraphics[width=\textwidth]{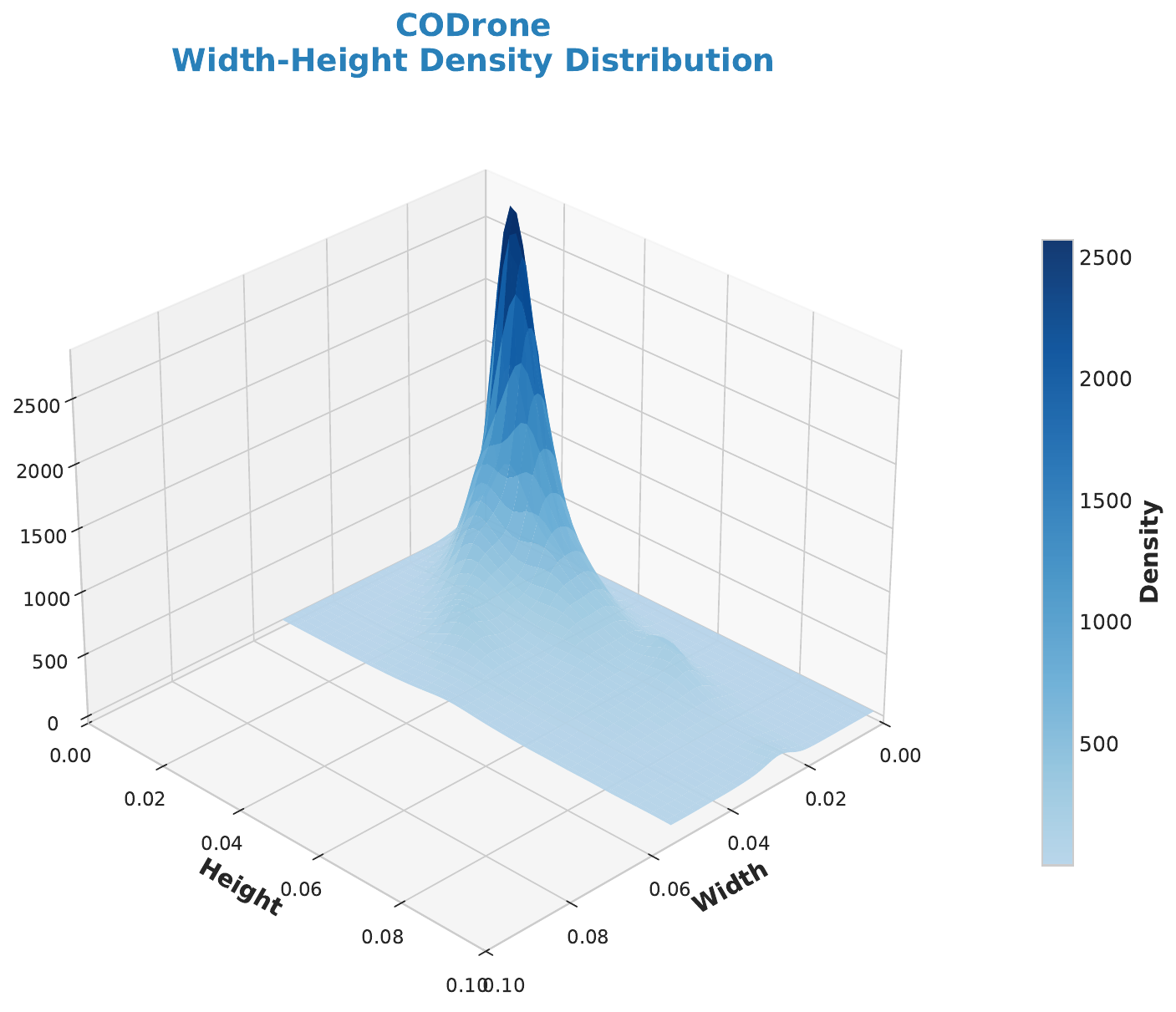}
        \caption*{\textbf{CODrone}}\vspace{-1mm}
      \end{subfigure}
    \caption*{(a)}
    \includegraphics[width=0.48\textwidth]{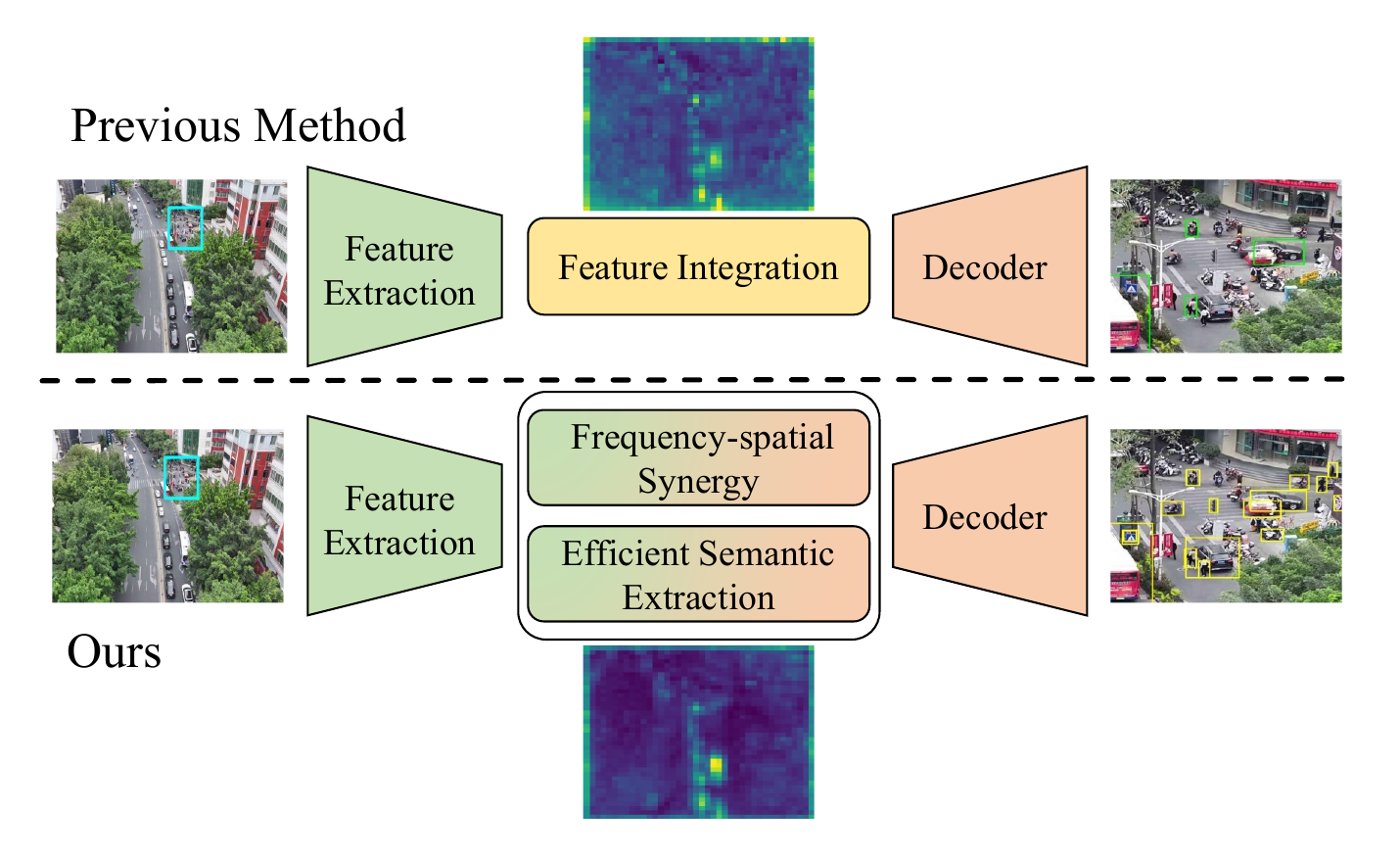}\vspace{-1mm}
    \caption*{(b)}
    \caption{(a) 3D density distribution of object width and height in the VisDrone and CODrone datasets, illustrating that a significant portion of objects are small and cover only a limited pixel area within the images. (b) Compared with previous methods, our method aims to combine efficient semantic feature enhancement with dynamic frequency-spatial guidance during feature integration phase.}
    \label{fig_1}
\end{figure}

%
%
%
%
\IEEEPARstart{U}{}nmanned Aerial Vehicle (UAV) imagery, characterized by its diverse perspectives, flexible acquisition, and wide-area coverage, has attracted increasing research attention\citep{zu1,zu2,tmm1,tmm2,tmm3,tmm4,tmm5,tmm6}, particularly in the field of object detection. More broadly, recent video-centric datasets for safety-critical urban perception, such as falling-object detection around buildings, further highlight the demand for robust object-level understanding in complex outdoor scenes\citep{tu2025fade}. Despite its potential, UAV imagery presents unique challenges for conventional object detection frameworks, primarily due to the prevalence of small objects with limited pixel representation as shown in Fig.~\ref{fig_1}(a). This insufficient pixel coverage severely restricts the ability of standard convolutional operations to extract discriminative features. Furthermore, the downsampling operations commonly employed in convolutional neural networks (CNNs) often lead to the loss of fine-grained details that are critical for accurate detection\citep{FPN}. Meanwhile, many existing methods rely on static fusion strategies that struggle to capture deep semantic representations effectively and fail to exploit crucial information in the frequency domain to preserve fine structural and textural details of small objects, thus limiting the detection performance in complex UAV scenarios\citep{rtdetr,yolov10,trans+}.

While YOLO-series detectors achieved remarkable success in generic object detection scenarios\citep{Scaled-YOLOv4,yolox,tian2025yolov12}, their performance tends to degrade significantly in UAV applications. A key limitation lies in their anchor-based design, which struggles to accommodate the extreme scale variations inherent in aerial imagery. Specifically, predefined anchor boxes often fail to cover the full range of object sizes, especially for small objects that occupy only a few pixels. Moreover, the single-stage detection paradigm adopted by YOLO prioritizes efficiency at the cost of representational flexibility. Direct regression from features to bounding boxes provides limited capacity to model complex spatial relationships and contextual cues, which are vital in cluttered aerial scenes. 

In contrast, the Real-Time Detection Transformer (RT-DETR) offers several architectural advantages that are well aligned with the unique demands of UAV imagery\citep{rtdetr}. Its end-to-end, anchor-free design eliminates the need for hand-crafted anchors and post-processing steps such as Non-Maximum Suppression (NMS), resulting in more flexible object localization and reduced inference complexity. The transformer-based backbone further enhances performance by capturing long-range dependencies and global context through self-attention mechanisms\citep{att}, which are essential for identifying dispersed objects in aerial views. Additionally, the decoder structure supports complex feature interactions, enabling more nuanced modeling of spatial relationships. Nevertheless, RT-DETR is not without limitations. While it addresses many of the architectural constraints present in YOLO, it remains suboptimal for UAV imagery. In particular, it lacks effective utilization of frequency information and exhibits limitations in extracting rich semantic features efficiently.

To address these issues, we propose \textbf{EFSI-DETR}, a novel detection framework that integrates efficient semantic feature enhancement with dynamic frequency-spatial guidance, as illustrated in Fig.\ref{fig_1}(b). Built upon the RT-DETR architecture, EFSI-DETR boosts it with exploration of frequency-aware processing and semantic feature extraction mechanisms. The key contributions of this work are as follows:
\begin{itemize}

    \item We introduce \textbf{EFSI-DETR}, an efficient detector designed for complex UAV scenarios, with \textbf{Fine-grained Feature Retention (FFR)} strategy that supplements semantics with detailed spatial cues. Our framework achieves superior accuracy while maintaining real-time performance. 
    
    \item We propose \textbf{Dynamic Frequency-spatial Unified Synergy Network (DyFusNet)}, a frequency-spatial fusion module, which decomposes features into multi-resolution spectral components in a simulated manner and adaptively integrates them based on the input characteristics.

    \item We exploit the \textbf{Efficient Semantic Feature Concentrator (ESFC)}, a lightweight and semantically-aware module, which enhances deep semantic feature extraction by leveraging content-adaptive convolution selection.


    \item We conduct comprehensive evaluations on two challenging UAV benchmarks \textit{i.e.} VisDrone and CODrone. Our EFSI-DETR achieves notable improvement in both accuracy and efficiency, surpassing the state-of-the-arts by \textbf{1.6}\% in AP and \textbf{5.8}\% in AP$_s$ on the VisDrone dataset, meanwhile maintaining a high inference speed of \textbf{188} FPS on one RTX 4090 GPU.
\end{itemize}

\section{Related Work}

\paragraph{Real-time Object Detection}
Real-time object detection has been a cornerstone of computer vision applications. Single-stage detectors such as the YOLO family\citep{redmon2016you} and FCOS\citep{tian2019fcos} have achieved remarkable performance by balancing accuracy and efficiency.  Recent advances include YOLO11, YOLOv12\citep{Jocher_Ultralytics_YOLO_2023,tian2025yolov12}, and RT-DETR\citep{rtdetr}, which have pushed the boundaries of real-time detection on standard benchmarks like COCO\citep{lin2014microsoft}. However, these methods primarily focus on natural images with moderate resolutions and struggle with UAV imagery captured at high altitudes where small objects are prevalent. 

\paragraph{Small Object Detection}
Small object detection presents unique challenges due to limited pixel representation and susceptibility to background noise. Related dense prediction studies also indicate that improving inter-class feature separability can benefit fine-grained pixel-level discrimination\citep{zhang2022distilling}, which is especially relevant when small targets provide weak visual evidence. Traditional approaches have focused on data augmentation\citep{kisantal2019augmentation} and multi-scale training\citep{singh2018analysis} to improve small object recognition. More sophisticated methods, like ClusDet\citep{yang2019clustdet}, which employs cluster-based scale estimation, and DM-Net\citep{li2020dmnet} which utilizes density maps for spatial context modeling. Recent works have explored attention mechanisms and feature enhancement techniques: QueryDet\citep{yang2022querydet} introduces query-based acceleration for pyramid features, and CEASC\citep{Du2023ceasc} applies context-enhanced sparse convolution for global information capture. Although these methods introduce lightweight and decoupled heads that provide certain acceleration benefits, achieving real-time object detection continues to pose a substantial challenge. 

\paragraph{Generative Priors for Visual Perception}
Recent diffusion-based models have also shown promise as visual priors beyond image generation. For example, text-to-image diffusion models can provide semantic and appearance priors for unsupervised visual object tracking\citep{zhang2026leveraging}. Although our work focuses on real-time detection rather than tracking or generation, these studies suggest that high-level semantic priors can complement local discriminative features, which is consistent with our goal of strengthening semantic feature concentration for UAV small-object detection.

\paragraph{Multi-scale Feature Fusion}
Due to limited pixel representation in feature maps, small objects often lack sufficient semantic and spatial cues. To alleviate this, multi-scale feature fusion has become a key strategy. Feature Pyramid Network (FPN)\citep{FPN} established the foundation by combining semantically rich deep features with spatially detailed shallow features. PANet\citep{liu2018path} introduced bidirectional fusion, while BiFPN\citep{tan2020efficientdet} employed weighted aggregation for efficiency. PAFPN\citep{liu2018path} added a bottom-up path, and NAS-FPN\citep{ghiasi2019nasfpn} leveraged neural architecture search for optimal fusion. Recent methods like FCOS\citep{tian2019fcos} and YOLOF\citep{chen2021you} attempted to simplify multi-scale detection, while maintaining performance. However, most of these methods rely on static fusion strategies, leading to redundant and inefficient extraction of deep semantic information. In addition, informative frequency-domain cues are crucial for small-object detection, and recent frequency-enhanced diffusion models further show that spectral cues can strengthen semantic alignment in visual representations\citep{zhou2026frequency}. Prior frequency-aware detectors like UAV-DETR\citep{uavdetr} typically operate explicitly in transform domain to extract band-limited features. Despite their expressiveness, these designs require global memory permutations and create non-fusible edges in computation graph, ultimately limiting kernel fusion and undermining inference efficiency on modern hardware.

In this work, we propose EFSI-DETR, a novel DETR-based framework to address existing limitations through efficient semantic feature enhancement with dynamic frequency-spatial guidance, thereby improving small object detection performance while preserving real-time efficiency.

\section{Method}
\begin{figure*}[!htbp]
\centering
\includegraphics[width=1\textwidth]{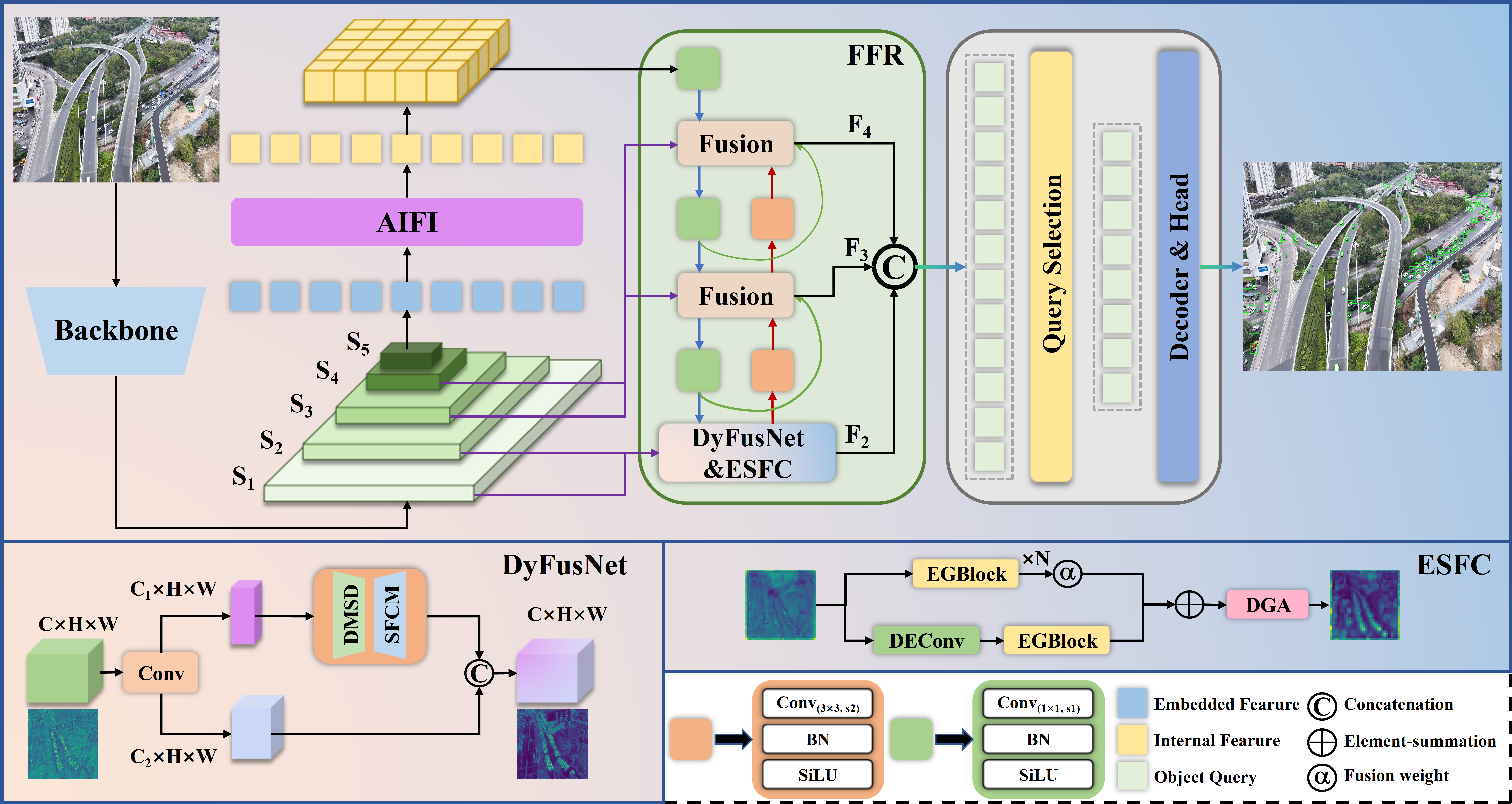} 
\caption{Overview of EFSI-DETR. DyFusNet enables robust and adaptive multi-scale feature fusion via Dynamic Multi-resolution Spectral Decomposition (DMSD) and Spatial-Frequency Cooperative Modulation (SFCM). ESFC enhances deep semantic representations via Dynamic Expert Convolution (DEConv), Efficient Ghost Block (EGBlock), and Dual-domain Guidance Aggregation (DGA). FFR incorporates spatially rich shallow features during the fusion process, effectively preserving fine-grained spatial details—crucial for small object detection. S$_1$-S$_5$ denote the shallow-to-deep feature stages of the backbone network. The encoder produces multi-scale fused features F$_2$–F$_4$, which are forwarded to the decoder for final prediction. Notably, the structures of the AIFI and Fusion blocks are consistent with those employed in RT-DETR\citep{rtdetr}.}
\label{fig11}
\end{figure*}

UAV-view imagery differs from common-view detection in two key aspects: (i) targets often occupy only a few pixels, making localization highly dependent on fine-grained spatial details, and (ii) complex aerial backgrounds introduce clutter and aliasing effects that suppress weak object cues. To explicitly address these UAV-specific challenges, the overall architecture of EFSI-DETR, as illustrated in Fig.~\ref{fig11}, is designed with three UAV-oriented components: Dynamic Frequency-Spatial Unified Synergy Network (DyFusNet), Efficient Semantic Feature Concentrator (ESFC), and Fine-grained Feature Retention (FFR). Unlike general multi-branch widening architectures, DyFusNet enables efficient multi-scale feature fusion via frequency-inspired, non-transform frequency-spatial guidance, thus retaining structural cues of tiny objects without introducing transform-related computational overhead. ESFC further enhances semantic discrimination through a lightweight semantic concentrator featuring low parameter complexity and content-adaptive semantic feature extraction, making it well-suited for deployment on UAV platforms. FFR explicitly preserves shallow, high-resolution features and prioritizes fine-grained spatial cues while discarding overly coarse deep features during fusion to improve tiny-object localization and reduce semantic redundancy. Collectively, these components deliberately designed to match the characteristics of UAV-view imagery, distinguish EFSI-DETR from traditional common-view object detection frameworks.
\subsection{Dynamic Frequency-spatial Unified Synergy Network}

The Dynamic Frequency-Spatial Unified Synergy Network introduces a novel paradigm for multi-scale feature representation learning by leveraging the complementary nature of frequency and spatial information with its pseudocode provided in Algorithm~\ref{alg:dyfusenet}. Unlike classical frequency-domain methods that rely on explicit transforms, DyFusNet adopts a frequency-inspired but non-FFT formulation to preserve deployment efficiency on modern inference engines (TensorRT/ONNX back-ends). Instead of transforming features into a complex spectral basis, DyFusNet constructs learnable low/mid/high-band proxies directly in the spatial domain using light-weight operators that are hardware-friendly. This choice avoids transform overhead, complex-to-real conversions, and poor kernel-fusion characteristics that often hinder real-time throughput.
\begin{algorithm}[!htbp]
\small
\caption{DyFusNet: Dynamic Frequency-spatial Unified Synergy Network}
\label{alg:dyfusenet}
\begin{algorithmic}[1]
\Require Input $\mathbf{X}\!\in\!\mathbb{R}^{B\times C\times H\times W}$; expand ratio $e\!\in\!(0,1)$
\Ensure Output $\mathbf{Y}\!\in\!\mathbb{R}^{B\times C\times H\times W}$
\State $C_{\text{hid}}\gets \lfloor C\cdot e\rfloor$
\State $\mathbf{F}\gets \textsc{Conv1x1}(\mathbf{X};\,C\!\to\!C)$ \Comment{reduce}
\State $[\mathbf{F}_1,\mathbf{F}_2]\gets \textsc{Split}\!\big(\textsc{Conv1x1}(\mathbf{F});\ [\,C_{\text{hid}},\, C{-}C_{\text{hid}}\,]\ \text{along channels}\big)$
\State $\mathbf{F}_1\gets \textsc{DMSD}(\mathbf{F}_1)$ \Comment{dynamic multi-resolution spectral decomposition}
\State $\mathbf{F}_1\gets \textsc{SFCM}(\mathbf{F}_1)$ \Comment{spatial--frequency cooperative modulation}
\State $\mathbf{Y}\gets \textsc{Conv1x1}\!\big(\textsc{Concat}(\mathbf{F}_1,\mathbf{F}_2);\ C\!\to\!C\big)$ \Comment{fuse}
\State \Return $\mathbf{Y}$
\Statex
\Function{DMSD}{$\mathbf{U}\!\in\!\mathbb{R}^{B\times C_{\text{hid}}\times H\times W}$}
  \State $\mathbf{v}\gets \textsc{GAP}(\mathbf{U})$ \Comment{$\mathbf{v}\!\in\!\mathbb{R}^{B\times C_{\text{hid}}\times 1\times 1}$}
  \State $\mathbf{w}\gets \textsc{Softmax}_c\!\big(\textsc{Conv1x1}(\mathbf{v};\,C_{\text{hid}}\!\to\!3)\big)$
  \State $\mathbf{L}\gets \textsc{AvgPool}_{3\times3,\,s=1,\,p=1}(\mathbf{U})$ \Comment{low-}
  \State $\mathbf{M}\gets \mathbf{U}$ \Comment{mid-}
  \State $\mathbf{H}\gets \textsc{DWConv}_{3\times3,\,s=1,\,p=1}(\mathbf{U})$ \Comment{high-}
  \State \Return $\mathbf{w}_{[:,0]}\!\odot\!\mathbf{L}\;+\;\mathbf{w}_{[:,1]}\!\odot\!\mathbf{M}\;+\;\mathbf{w}_{[:,2]}\!\odot\!\mathbf{H}$
\EndFunction
\Statex
\Function{SFCM}{$\mathbf{U}\!\in\!\mathbb{R}^{B\times C_{\text{hid}}\times H\times W}$}
  \State $\mathbf{A}\gets \textsc{Conv1x1}(\mathbf{U};\,C_{\text{hid}}\!\to\!C_{\text{hid}})$
  \State $\mathbf{B}\gets \textsc{DWConv}_{3\times3,\,p=1}(\mathbf{U})$ \Comment{local $3{\times}3$}
  \State $\mathbf{C}\gets \textsc{DWConv}_{5\times5,\,p=2}(\mathbf{U})$ \Comment{wider $5{\times}5$}
  \State $\mathbf{F}\gets \mathbf{A}+\mathbf{B}+\mathbf{C}$ \Comment{spatial--frequency fusion}
  \State $\mathbf{s}\gets \textsc{GAP}(\mathbf{F})$ \Comment{$\mathbf{s}\!\in\!\mathbb{R}^{B\times C_{\text{hid}}\times 1\times 1}$}
  \State $\mathbf{t}\gets \textsc{ReLU}\!\big(\textsc{Conv1x1}(\mathbf{s};\,C_{\text{hid}}\!\to\!C_{\text{hid}}/4)\big)$
  \State $\mathbf{a}\gets \sigma\!\big(\textsc{Conv1x1}(\mathbf{t};\,C_{\text{hid}}/4\!\to\!C_{\text{hid}})\big)$
  \State \Return $\mathbf{F}\odot \mathbf{a}$
\EndFunction
\end{algorithmic}
\end{algorithm}
\paragraph{Dynamic Multi-resolution Spectral Decomposition} Rather than employing fixed frequency decomposition, the Dynamic Multi-resolution Spectral Decomposition (DMSD) module introduces a learnable mechanism that adaptively weights different frequency-inspired components based on input content. Notably, DMSD is not designed to reproduce a discrete wavelet transform or its oriented sub-bands in a strict signal-processing sense; instead, it provides a deployable, spatial-domain approximation that emphasizes different spectral tendencies through lightweight operators.

The module decomposes input features into three complementary pathways to simulate different frequency responses:
\begin{equation}
\mathcal{F}_{\mathrm{DMSD}}(\mathbf{X}) = \sum_{i \in \{\text{low},\text{mid},\text{high}\}} \alpha_i(\mathbf{X})\, \mathcal{H}_i(\mathbf{X})
\label{eq:DMSD_main}
\end{equation}
where $\mathcal{H}_i$ represents frequency-selective operators and $\alpha_i(\mathbf{X})$ denotes content-adaptive weights that dynamically modulate the contribution of each frequency component based on the content distribution. The frequency-selective operators are designed to capture complementary spectral characteristics:
\begin{equation}
[\alpha_{\text{low}},\,\alpha_{\text{mid}},\,\alpha_{\text{high}}]^{\top}
=\mathrm{softmax}\!(\mathbf{W}_2\,\cdot\phi\!(\mathbf{W}_1\,\cdot\mathrm{GAP}(\mathbf{X}))),
\end{equation}
\begin{equation}
\begin{aligned}
\mathcal{H}_\text{low}(\mathbf{X}) &= \text{AvgPool}_{3 \times 3}(\mathbf{X})\\
\mathcal{H}_\text{mid}(\mathbf{X}) &= \text{Identity}(\mathbf{X})\\
\mathcal{H}_\text{high}(\mathbf{X}) &= \text{Conv}_{3 \times 3}^{dw}(\mathbf{X})
\end{aligned}
\end{equation}
where $\mathrm{GAP}$ and $\phi$ denote global average pooling and GELU. The low-frequency component $\mathcal{H}_\text{low}$ employs average pooling to capture smooth, global structural information, analogous to low-pass filtering. The mid-frequency component $\mathcal{H}_\text{mid}$ preserves the original features to preserve detailed information. The high-frequency component $\mathcal{H}_\text{high}$ utilizes depthwise convolution to enhance edge and texture details, effectively acting as a learnable high-pass filter adapts to input characteristics. Consequently, the three pathways should be interpreted as frequency-selective proxies rather than wavelet coefficients.

This design allows the module to adaptively emphasize different frequency components, such as enhancing high-frequency details in texture-rich regions while accentuating low-frequency components in smoother background areas.

\textbf{Why simulated frequency processing.}
In natural image analysis, frequency selectivity can be achieved either by explicit transforms or by local, learnable spatial operators whose impulse responses approximate canonical low-, band-, or high-pass filters. Since convolution in space corresponds to multiplication in the Fourier domain, local kernels inherently induce frequency responses. Average pooling behaves as a strong low-pass operator, the identity path is all-pass, and depthwise convolutions can learn derivative-like kernels that emphasize high-frequency edges and textures. In this view, DMSD forms a compact, learnable filter-bank that is frequency-inspired yet intentionally non-transform: it does not enforce the orthogonality or orientation constraints of DWT sub-bands, but instead pursues content-adaptive band emphasis with minimal overhead. This forms a compact filter-bank view consistent with multiresolution analysis\citep{mallat89theory} and spatial architectures that explicitly separate frequency pathways\citep{chen2019octconv, zhang2019blurpool}. Unlike fixed banks, our dynamic weighting $\alpha_i(\mathbf{X})$ adapts band emphasis to content: texture-rich regions receive stronger high-band responses, while smooth backgrounds favor the low band. Such simulated frequency processing preserves translation equivariance under standard padding, reduces aliasing with pre-filtering, avoids periodic-boundary artifacts from global transforms, and integrates naturally with modern detection backbones.

\textbf{Why not FFT-like transforms.}
Spectral transforms (FFT/DCT/DWT) have shown strong global-mixing ability in recognition models\citep{rippel2015spectral, lee-thorp2021fnet, worrall2017harmonic}, but they incur practical costs for real-time UAV deployment. (i) \textbf{Kernel fusion.} FFT pipelines—forward transform, pointwise multiply, inverse transform—are difficult to fuse with adjacent operations, causing additional kernel launches and synchronization. Local depthwise or pointwise ops are compiler-friendly and easily fused. (ii) \textbf{Memory traffic.} Global transforms require non-local data movement and complex-valued tensors, increasing bandwidth pressure. (iii) \textbf{Shape sensitivity.} FFT efficiency depends on input factorization; irregular sizes need padding or tiling, which complicates dense prediction near boundaries. (iv) \textbf{Hardware portability.} Highly optimized, int8-capable FFT kernels are less common across edge NPUs and embedded GPUs than standard convolution and pooling. Empirically, Fourier-based mixing enlarges receptive fields effectively but offers limited benefit under strict latency and memory budgets. We therefore adopt a deployable, learnable, and content-adaptive spatial approximation that retains key spectral advantages while ensuring efficient, stable inference.

\paragraph{Spatial–Frequency Cooperative Modulation}
To jointly capture multi–receptive-field structure and channel-wise discriminability, we integrate spatial aggregation and frequency-guided channel reweighting into a single module, termed Spatial–Frequency Cooperative Modulation (SFCM). Given an input feature map $\mathbf{X}\in\mathbb{R}^{C\times H\times W}$, SFCM first aggregates spatial evidence using a parallel multi-kernel operator that discretizes scale-space responses:
\begin{align}
\mathbf{Z}(\mathbf{X})&=\underbrace{\mathbf{W}_{1\times1}\cdot\mathbf{X}}_{\text{channel mixing}}+\sum_{k\in\{3,5\}}\underbrace{\mathbf{W}^{\text{dw}}_{k}\cdot\mathbf{X}}_{\text{depthwise }k\times k},\label{eq:sfcm_spatial}
\end{align}
where $\mathbf{W}_{1\times1}\!\in\!\mathbb{R}^{C\times C\times1\times1}$ denotes a point-wise convolution for cross-channel mixing and $\mathbf{W}^{\text{dw}}_{k}\!\in\!\mathbb{R}^{C\times1\times k\times k}$ are depthwise convolutions with kernel size $k$. This additive synthesis preserves fine structures (via $3\times 3$), aggregates broader context (via $5\times 5$), and couples them through channel projection (via $1\times 1$), yielding a spatially enriched tensor $\mathbf{Z}(\mathbf{X})\in\mathbb{R}^{C\times H\times W}$ suitable for subsequent modulation.
Channel modulation is then applied to $\mathbf{Z}(\mathbf{X})$ by estimating per-channel gates from its global statistics. We form a channel descriptor $\mathbf{s}\in\mathbb{R}^{C}$ and map it to attention coefficients $\boldsymbol{\beta}\in(0,1)^C$ by a two-layer bottleneck with a point-wise nonlinearity:
\begin{align}
\mathbf{s}&=\mathrm{GAP}(\mathbf{Z}(\mathbf{X})),\label{eq:sfcm_gap}\\
\boldsymbol{\beta}(\mathbf{X})&=\sigma\!\left(\mathbf{W}_2\,\cdot\delta\!\left(\mathbf{W}_1\,\cdot\mathbf{s}\right)\right),\label{eq:sfcm_att}
\end{align}
where $\delta$ is a ReLU and $\sigma$ is a sigmoid applied channel-wise. $\mathbf{W}_1\in\mathbb{R}^{\tfrac{C}{r}\times C}$ and $ \mathbf{W}_2\in\mathbb{R}^{C\times \tfrac{C}{r}}$ represent convolution weights. $r$=4 is the reduction ratio. The final cooperative modulation multiplies the spatially aggregated tensor with the attention vector broadcast along spatial dimensions:
\begin{align}
\mathcal{F}_{\mathrm{SFCM}}(\mathbf{X})&=\mathbf{Z}(\mathbf{X})\odot \boldsymbol{\beta}(\mathbf{X}).\label{eq:sfcm_final}
\end{align}

In UAV scenes with dense backgrounds and small objects, $\mathcal{F}_{\mathrm{SFCM}}$ preserves high-frequency details captured by depthwise branches while adaptively emphasizing channels that carry discriminative responses and suppressing channels dominated by clutter or noise.

DMSD and SFCM are used inside the frequency path of DyFusNet. Let $\mathbf{X}\in\mathbb{R}^{C\times H\times W}$ be the backbone input. We partition channels into two subsets to avoid uniformly applying spectral–spatial processing to all channels:
\begin{align}
\mathbf{X}_1,\mathbf{X}_2&=\mathrm{Split}(\mathbf{X},\text{ratio}=e),\label{eq:dyfus_split}
\end{align}
where $e$ is set to 0.25 to control the proportion routed to the frequency path. The frequency path first performs dynamic multi-resolution spectral decomposition and then applies the proposed cooperative modulation:
\begin{align}
\mathcal{F}_{\mathrm{freq}}&=\mathcal{F}_{\mathrm{SFCM}}\circ \mathcal{F}_{\mathrm{DMSD}},\label{eq:dyfus_freq}
\end{align}
and the unified output is obtained by concatenation followed by a $1\times 1$ fusion:
\begin{align}
\mathcal{F}_{\mathrm{DyFusNet}}(\mathbf{X})&=\mathrm{Concat}\!\left(\mathcal{F}_{\mathrm{freq}}(\mathbf{X}_1),\,\mathbf{X}_2\right).\label{eq:dyfus_out}
\end{align}
In practice, the channel descriptor in~\eqref{eq:sfcm_att} is computed on $\mathbf{Z}(\mathbf{X}_1)$, making the gates responsive to the spatially aggregated, band-emphasized activations delivered by $\mathcal{F}_{\mathrm{DMSD}}$. The resulting pipeline consolidates multi-scale spatial evidence and frequency-guided channel selection without resorting to explicit Fourier operators, which is favorable for optimized inference backends and deployment-critical UAV applications.

\subsection{Efficient Semantic Feature Concentrator}

As shown in Fig.~\ref{fig11}, the proposed ESFC module employs a dual-branch architecture with learnable fusion weights, striking an effective balance between computational efficiency and representational capacity, with its pseudocode detailed in Algorithm~\ref{alg:esfc}. This design significantly enhances semantic feature extraction, which is critical for real-time detection of small objects in UAV imagery.

\begin{algorithm}[!htbp]
\small
\caption{ESFC: Efficient Semantic Feature Concentrator}
\label{alg:esfc}
\begin{algorithmic}[1]
\Require Input $\mathbf{X}\in\mathbb{R}^{B\times C_1\times H\times W}$; output channels $C_2$; depth $n$; DGA params $(\gamma,b,k_s)$; learnable $\alpha$
\Ensure Output $\mathbf{Y}\in\mathbb{R}^{B\times C_2\times H\times W}$

\State $\mathbf{X}_1 \gets \textsc{DEConv}(\mathbf{X};\, C_1\!\to\! C_1, k{=}1, s{=}1, E{=}3)$
\For{$i=1$ to $n$}
  \State $\mathbf{X}_1 \gets \textsc{EGBlock}(\mathbf{X}_1;\, C{=}C_1)$
\EndFor
\State $\mathbf{X}_2 \gets \textsc{EGBlock}(\mathbf{X};\, C{=}C_1)$ \Comment{shortcut}
\State $\mathbf{F} \gets \mathbf{X}_1 + \alpha\cdot \mathbf{X}_2$
\State $\mathbf{F} \gets \textsc{DGA}(\mathbf{F};\, \gamma,b,k_s)$
\If{$C_1 \neq C_2$}
    \State $\mathbf{Y} \gets \textsc{Conv1x1BNAct}(\mathbf{F};\,C_1\!\to\!C_2)$
\Else
    \State $\mathbf{Y} \gets \mathbf{F}$
\EndIf
\State \Return $\mathbf{Y}$

\Statex
\Function{DEConv}{$\mathbf{U};\, C_{\text{in}}\!\to\!C_{\text{out}},\, k, s, E$}
  \State $\mathbf{v}\gets \textsc{GAP}(\mathbf{U})$
  \State $\mathbf{a}\gets \textsc{Softmax}\big(\textsc{Conv1x1}(\mathbf{v});\,\text{dim}{=}E\big)$ 
        \Comment{$\mathbf{a}\in\mathbb{R}^{B\times E\times 1\times 1}$}
  \For{$i=1$ to $E$}
    \State $\mathbf{O}_i \gets \textsc{Conv2D}_i(\mathbf{U};\,k, s, \text{pad}{=}k/2)$
  \EndFor
  \State $\mathbf{O}\gets \sum_{i=1}^{E} \mathbf{O}_i \odot \mathbf{a}_{[:,i,:,:]}$
  \State \Return $\textsc{SiLU}\!\big(\textsc{BN}(\mathbf{O})\big)$
\EndFunction

\Statex
\Function{EGBlock}{$\mathbf{U};\, C$} 
  \State $C_h \gets \lfloor C/2 \rfloor$
  \State $\mathbf{P}\gets \textsc{SiLU}\!\big(\textsc{BN}(\textsc{Conv1x1}(\mathbf{U};\,C_h))\big)$
  \State $\mathbf{C}_{\text{cheap}}\gets \textsc{DWConv3x3}(\mathbf{P};\,\text{groups}{=}C_h)$
  \State $\mathbf{C}_{\text{cheap}}\gets \textsc{SiLU}(\textsc{BN}(\mathbf{C}_{\text{cheap}}))$
  \State $\mathbf{Z}\gets \textsc{Concat}(\mathbf{P},\,\mathbf{C}_{\text{cheap}})$ \Comment{$\mathbf{Z}\in\mathbb{R}^{B\times C\times H\times W}$}
  \State \Return $\mathbf{Z}$
\EndFunction

\Statex
\Function{DGA}{$\mathbf{U}\in\mathbb{R}^{B\times C\times H\times W};\,\gamma,b,k_s$}
  \State $t\gets \big\lfloor \big|\tfrac{\log_2(C)+b}{\gamma}\big| \big\rfloor$;\quad $k_c\gets t$ if $t$ odd else $t{+}1$
  \State $\mathbf{z}\gets \textsc{GAP}(\mathbf{U})\in\mathbb{R}^{B\times C\times 1\times 1}$
  \State $\mathbf{s}_c\gets \sigma\!\big(\textsc{Conv1D}(\mathbf{z};\,\text{kernel}{=}k_c,\ \text{pad}{=}k_c/2)\big)$
  \State $\mathbf{U}_c\gets \mathbf{U}\odot \textsc{Reshape}(\mathbf{s}_c, B{\times}C{\times}1{\times}1)$
  \State $\mathbf{u}_{\text{avg}}\gets \textsc{Mean}(\mathbf{U}_c,\ \text{dim}{=}C,\ \text{keepdim}{=}1)$
  \State $\mathbf{u}_{\text{max}}\gets \textsc{Max}(\mathbf{U}_c,\ \text{dim}{=}C,\ \text{keepdim}{=}1)$
  \State $\mathbf{U}_s\gets \textsc{Concat}[\mathbf{u}_{\text{avg}},\mathbf{u}_{\text{max}}]$
  \State $\mathbf{a}_s\gets \sigma\!\big(\textsc{Conv2D}(\mathbf{U}_s;\,\text{in}{=}2,\ \text{out}{=}1,\ \text{kernel}{=}k_s,\ \text{pad}{=}k_s/2)\big)$
  \State \Return $\mathbf{U}_c\odot \mathbf{a}_s$
\EndFunction
\end{algorithmic}
\end{algorithm}

\paragraph{Dynamic Expert Convolution}

Traditional convolutional operations employ static kernels that may not optimally adapt to varying feature distributions across different spatial regions. To address this, we propose Dynamic Expert Convolution (DEConv), which leverages multiple expert convolutions with learned attention weights for adaptive kernel selection.

DEConv employs $K$ expert convolutions $\mathbf{W}_k$ with learned attention weights $\delta_k$:

\begin{equation}
\mathcal{F}_{DEConv}(\mathbf{X}) = \sum_{k=1}^{K} \delta_k  \mathbf{W}_k \cdot \mathbf{X}
\end{equation}

\paragraph{Efficient Ghost Block}
To address the computational overhead introduced by DEConv, ESFC integrates the Efficient Ghost Block (EGBlock) inspired by the ghost convolution principle\citep{ghostnet}, which provides a more efficient alternative with reduced redundancy.
\begin{equation}
\begin{aligned}
\mathbf{F}_{primary} &= \Phi(\mathbf{W}_{primary} \cdot \mathbf{X}) \\
\mathbf{F}_{ghost} &= \Phi(\mathbf{W}_{cheap} \cdot \mathbf{F}_{primary}) \\
\mathcal{F}_{EGBlock}(\mathbf{X}) &= \text{Concat}(\mathbf{F}_{primary}, \mathbf{F}_{ghost})
\end{aligned}
\end{equation}

where $\mathbf{W}_{primary}$ is a standard convolution with reduced channels, $\mathbf{W}_{cheap}$ is a parameter-efficient depthwise convolution, and $\Phi$ represents the activation function. This design significantly reduces computational complexity while maintaining feature representation capability.


In addition, we incorporate a residual pathway composed of $N$ EGBlocks as shown in Fig.~\ref{fig11}, which ensures both robust feature extraction and training convergence.

\paragraph{Dual-domain Guidance Aggregation}

The Dual-domain Guidance Aggregation (DGA) module implements a sophisticated guidance that operates in both channel and spatial domains to enhance feature discriminability. The channel guidance $\mathbf{G}_{c}$ employs an adaptive kernel size strategy based on ECA-Net\citep{eca}:

\begin{equation}
\begin{aligned}
k = \psi(\log_2(C)) = \left|\frac{\log_2(C) + b}{\gamma}\right|_{\text{odd}}\\
\mathbf{G}_{c}(\mathbf{X}) = \mathbf{W}_{k}\cdot\text{AvgPool}(\mathbf{X})
\end{aligned}
\end{equation}

where $C$ is the number of channels, hyperparameters $b$=1 and $\gamma$=2 are configured in accordance with the original settings of ECA-Net, and $|\cdot|_{\text{odd}}$ ensures an odd kernel size for proper padding. This adaptive mechanism ensures optimal receptive field coverage for different channel dimensions.

The spatial guidance $\mathbf{G}_{s}$ uses a lightweight implementation that aggregates both average and max pooling features:

\begin{equation}
\mathbf{G}_{s}(\mathbf{X}) = \sigma(\mathbf{W}_{s}\cdot \text{Concat}(\text{AvgPool}(\mathbf{X}), \text{MaxPool}(\mathbf{X})))
\end{equation}

where $\sigma$ denotes the sigmoid activation and $\mathbf{W}_{s}$ is a single convolution layer. It preserves spatial attention effectiveness while minimizing the computational overhead. DGA enhances feature through cascaded channel and spatial guidance:
\begin{equation}
\mathcal{F}_{DGA}(\mathbf{X}) = \mathbf{G}_{s}(\mathbf{G}_{c}(\mathbf{X}))
\end{equation}

\subsection{Fine-grained Feature Retention}
Aerial imagery captured by UAVs typically contains numerous small-scale objects that occupy only a few pixels, which are easily obscured by background clutter. Intuitively, the successful detection of small objects is highly dependent on the finer-grained, high-resolution features captured in the early stages of the backbone. Deeper layers encode stronger semantic cues, the reduced spatial resolution often results in losing fine details critical for small object localization.

To address this issue, we refine the HybridEncoder\citep{rtdetr} within the RT-DETR framework by integrating low-level feature map \textbf{S}$_1$ and \textbf{S}$_2$, which preserves fine-grained spatial details. For the decoder, we deliberately exclude the coarse high-level semantic feature map \textbf{F}$_5$ and instead emphasize the use of \textbf{F}$_2$, \textbf{F}$_3$, and \textbf{F}$_4$. This design prioritizes high-resolution features that maintain spatial detail crucial for detecting small objects, while simultaneously reducing semantic redundancy and improving computational efficiency. 

\begin{table*}[ht]
\centering
\caption{Comparison in terms of AP (\%), latency, and parameters on VisDrone. ``\dag'' indicates that the result is directly taken from the original paper. ``*'' denotes that the input resolution exceeds 640$\times$640. ``--'' indicates that the result is not reported.}
\resizebox{\textwidth}{!}{
\begin{tabular}{l|c|cccccc|cc}
\toprule

\textbf{Model} & \textbf{Image Size} & \textbf{AP$^{val}$} & \textbf{AP$^{val}_{50}$} & \textbf{AP$^{val}_{75}$} & \textbf{AP$^{val}_s$} & \textbf{AP$^{val}_m$} & \textbf{AP$^{val}_l$} & \textbf{Latency (ms)} & \textbf{Params (M)} \\
\midrule
QueryDet$^\dag_{*}$~\citep{yang2022querydet}      & 800& 19.6 & 35.7 & 19.0 & -    & -    & -    & 288  & -   \\
RetinaNet$^\dag_{*}$~\citep{Lin2017RetinaNet}     & 800& 20.2 & 36.9 & 19.5 & -    & -    & -    & 14.7 & - \\
Faster-RCNN$^\dag_{*}$~\citep{Ren2015FasterRCNN}   & 800& 21.4 & 40.7 & 19.9 & 11.7 & 33.9 & 54.7 & 21.2 & - \\
CenterNet$^\dag_{*}$~\citep{centernet}     & 800& 27.8 & 47.9 & 27.6 & 21.3 & 42.1 & 49.8 & 95.2 & - \\
HRDNet$^\dag_{*}$~\citep{HRDNet}        & 1333& 28.3 & 49.3 & 28.2 & -    & -    & -    & -    & - \\
GFLV1$^\dag_{*}$~\citep{Li2020GFL}         & 1333& 28.4 & 50.0 & 27.8 & -    & -    & -    & 525  & -   \\
CEASC$^\dag_{*}$~\citep{Du2023ceasc}         & 1333& 28.7 & 50.7 & 28.4 & -    & -    & -    & 43.8 & - \\
RTMDet-L$^\dag$~\citep{Lyu2022RTMDet}      & 640& 23.7 & 37.4 & 25.5 & 12.5 & 38.7 & 50.4 & 13.9 & 52.3 \\
RemDet-M$^\dag$~\citep{remdet}      & 640& 28.2 & 46.1 & 28.9 & 18.2 & 41.7 & 51.0 & -  & 23.3 \\
RemDet-L$^\dag$~\citep{remdet}      & 640& 29.3 & 47.4 & 30.3 & 18.7 & 43.4 & \textbf{55.8} & 7.1  & 35.3 \\
\midrule
YOLOv8-L~\citep{Jocher_Ultralytics_YOLO_2023}   & 640 & 26.5 & 43.4 & 27.0 & 16.1 & 39.5 & 46.1 & 4.2 & 43.6 \\
YOLOv8-X~\citep{Jocher_Ultralytics_YOLO_2023}   & 640& 27.2 & 44.6 & 27.7 & 17.3 & 39.8 & 46.3 & 4.9 & 68.2\\
YOLOv10-L~\citep{yolov10}  & 640& 26.1 & 42.8 & 26.8 & 16.4 & 39.2 & 46.1 & 3.8 & 24.3 \\
YOLOv10-X~\citep{yolov10}  & 640& 27.2 & 44.5 & 28.0 & 17.4 & 40.3 & 48.3 & 4.5 & 29.4 \\
YOLO11-L~\citep{yolo11_ultralytics}   & 640& 27.0 & 44.0 & 27.6 & 16.5 & 40.4 & 51.6 & 4.2 & 25.3 \\
YOLO11-X~\citep{yolo11_ultralytics}   & 640& 27.7 & 45.1 & 28.4 & 18.1 & 40.4 & 47.6 & 5.8 & 56.8 \\
YOLOv12-L~\citep{tian2025yolov12}  & 640& 26.3 & 43.2 & 27.2 & 16.6 & 38.8 & 45.5 & 5.4 & 26.4 \\
YOLOv12-X~\citep{tian2025yolov12}   & 640& 28.1 & 45.7 & 29.1 & 17.9 & 41.5 & 46.5 & 6.5 & 59.1 \\
\midrule
RT-DETR-R18~\citep{rtdetr}       & 640& 25.7 & 43.3 & 25.9 & 17.5 & 35.8 & 39.4 & 3.6  & 19.9   \\
RT-DETR-R50~\citep{rtdetr}       & 640& 26.8 & 44.8 & 26.9 & 18.4 & 37.1 & 43.9 & 5.1 & 42.0   \\
DEIM-RT-DETRv2-R18~\citep{huang2024deim}         &  640& 27.1 & 45.7 & 27.1 & 18.8 & 37.1 & 46.8 & 1.2 & 19.9  \\
DEIM-RT-DETRv2-R50~\citep{huang2024deim}         &  640& 28.2 & 47.7 & 28.1 & 19.0 & 39.2 & 47.9 & 1.9 & 42.0   \\
UAV-DETR-R18$^\dag$~\citep{uavdetr}                 &  640& 29.8 & 48.8 & - & - & - & - & - & 20.0   \\
UAV-DETR-R50$^\dag$~\citep{uavdetr}                 &  640& 31.5 & 51.1 & - & - & - & - & - & 42.0   \\
\midrule
\textbf{EFSI-DETR} & 640 & \textbf{33.1} & \textbf{52.7} & \textbf{34.7} & \textbf{24.8} & \textbf{44.0} & 44.0 &  5.3 & 27.3 \\
\textbf{EFSI-DETR$^{*}$}  & 800 & \textbf{35.0} & \textbf{55.2} & \textbf{36.9} & \textbf{27.3} & \textbf{45.3} & 43.3 &  5.3 & 27.3 \\
\bottomrule
\end{tabular}
}
\label{tab_visdrone}
\end{table*}

\section{Experiments}
\subsection{Implementation Details}
To evaluate the effectiveness of our method, we conduct experiments on two public UAV benchmarks: VisDrone~\citep{visdrone} and CODrone~\citep{codrone}. The VisDrone dataset contains 8,599 images captured by UAV platforms at various locations and altitudes. It provides over 54,000 annotated bounding boxes spanning ten predefined object categories. The dataset is split into 6,471 training images, 548 validation images, and 1,580 testing images. These subsets are collected from different locations but under similar environmental conditions, all with a resolution of 2,000 $\times$ 1,500 pixels. The CODrone dataset consists of 5,002 training images, 3,002 validation images, and 2,000 testing images, with a resolution of 3,840 $\times$ 2,160 pixels across 12 object categories. Since CODrone only provides oriented bounding boxes, we convert them into axis-aligned bounding boxes by computing the minimum enclosing rectangles to facilitate performance evaluation.

Following previous works, all models are evaluated on the validation sets to ensure fair comparisons. Experiments are performed on two RTX 4090 GPUs. Models are trained for 300 epochs with a batch size of 8 using the AdamW optimizer, with an initial learning rate of 1 $\times$ 10$^{-4}$ and a momentum of 0.9. We evaluate the performance of our model using standard COCO-style evaluation metrics, including \textbf{AP} (computed with IoU thresholds spanning 0.5 to 0.95), \textbf{AP$_{50}$}, and \textbf{AP$_{75}$}, as well as \textbf{AP$_s$}, \textbf{AP$_m$}, and \textbf{AP$_l$}. Specifically, \textbf{AP$_{50}$}, and \textbf{AP$_{75}$} correspond to average precision at IoU thresholds of 0.5 and 0.75, respectively, while \textbf{AP$_s$}, \textbf{AP$_m$}, and \textbf{AP$_l$} are computed by stratifying objects according to their bounding-box area following the COCO convention: small ($\textless$32$^2$ px$^2$), medium ([32$^2$,96$^2$] px$^2$), and large ($\textgreater$96$^2$ px$^2$). The input images are uniformly resized to 640 × 640. 

Unless otherwise specified, EFSI-DETR adopts a ResNet-18 backbone by default in both the main experiments and ablation studies to achieve a favorable accuracy–efficiency trade-off. To ensure fair comparisons, we follow a unified evaluation protocol covering both training and deployment. Specifically, all YOLO-family baselines are trained using their official hyperparameters and training recipes, while all DETR-family baselines are retrained under a single unified hyperparameter configuration identical to ours. Methods annotated with “†” indicate results quoted directly from the original papers or official repositories without retraining, and are explicitly marked to avoid ambiguous interpretations. In addition, the subscript “*” denotes models trained with an input resolution larger than 640×640.
For latency evaluation, all exportable models are assessed under a unified deployment configuration: TensorRT FP16 inference on a single RTX 4090 GPU with a batch size of 1. For baseline methods marked with "†", if TensorRT-based latency measurements are unavailable, we adopt the results reported in their original papers or official repositories; such cases are explicitly annotated to preclude unfair inference speed comparisons.


\subsection{Energy and Deployment Considerations}
Energy efficiency is important for UAV deployment. In this work,  we provide deployment-oriented efficiency proxies that are reproducible and closely related to energy consumption under a fixed device setting, including TensorRT FP16 latency and model parameters. Following our unified benchmark protocol (input 640 × 640; TensorRT FP16 on a single RTX 4090), EFSI-DETR achieves real-time inference with low latency and moderate model size as quantified in Table~\ref{tab_visdrone}.

From a design perspective, EFSI-DETR is energy-aware by construction. Specifically, DyFusNet adopts a frequency-inspired yet spatial-domain formulation and avoids explicit FFT/DWT-like transforms, which often introduce additional kernel launches and non-local memory traffic that are less favorable for kernel fusion on modern inference backends. In contrast, our module primarily relies on hardware-friendly operators that are well supported by mainstream inference engines, making  overall pipeline more suitable for deployment-critical UAV scenarios.

\subsection{Comparison with SOTA on UAV Datasets}


\paragraph{Comparison Results on VisDrone}
Table~\ref{tab_visdrone} presents a comprehensive comparison between the proposed EFSI-DETR and recent state-of-the-art object detectors on the VisDrone dataset. The EFSI-DETR achieves a leading AP of 33.1\%, outperforming all compared methods. In particular, it exceeds the the strongest YOLO series model, YOLOv12-X\citep{tian2025yolov12}, by 5.0\% in AP and demonstrates a notable improvement of 7.0\% in AP$_{50}$. This advantage is especially pronounced for small objects, where EFSI-DETR achieves an AP$_s$ of 24.8\%, significantly higher than the 17.9\% obtained by YOLOv12-X. Such improvement is crucial for UAV-based detection where small-scale targets are frequent.

In comparison with detectors specifically designed for small object detection, such as RemDet-L\citep{remdet}, the proposed method also demonstrates superior performance and efficiency. EFSI-DETR improves AP$_s$ by 6.1\%, while also achieving gains of 3.8\% in AP and 5.3\% in AP$_{50}$. Furthermore, EFSI-DETR offers faster inference speed with a latency of 5.3\,ms, compared to 7.1\,ms for RemDet-L, and utilizes only 77.3\% of the parameter count, with 27.3M parameters versus 35.3M.

Additional comparisons with recent DETR-based models further highlight the effectiveness of the proposed design. EFSI-DETR surpasses DEIM-RT-DETRv2-R50\citep{huang2024deim} by 4.9\% in AP and 5.0\% in AP$_{50}$, demonstrating substantial improvements in small object detection with AP$_s$ exceeding by 5.8\%. With only 65.0\% of the parameters of DEIM-RT-DETRv2-R50, EFSI-DETR achieves competitive efficiency and outperforms in most key metrics.

While our method exhibits relatively limited performance on large objects, potentially due to the representational constraints inherent in the parameter-efficient architecture when capturing fine-grained details, this trade-off remains well justified. The proposed design achieves substantial improvements in small object detection precision while maintaining real-time inference speed, which aligns closely with the core requirements of UAV-based object detection scenarios. In future work, we plan to further investigate adaptive multi-scale fusion mechanisms to alleviate the degradation on large objects, aiming to achieve a more balanced detection performance across different object scales without sacrificing efficiency.

\begin{table}[ht]
\centering
\small 
\setlength{\tabcolsep}{1.1pt} 
\caption{Comparison in terms of AP (\%) on CODrone.}
\label{tab_codrone}
\begin{tabular}{l|cccccc}
\toprule
\textbf{Model} & \textbf{AP$^{val}$} & \textbf{AP$^{val}_{50}$} & \textbf{AP$^{val}_{75}$} & \textbf{AP$^{val}_s$} & \textbf{AP$^{val}_m$} & \textbf{AP$^{val}_l$}  \\
\midrule
YOLOv8-L~\citep{Jocher_Ultralytics_YOLO_2023}      & 16.3 & 30.7 & 15.7 & 1.3 & 10.8 & 28.5 \\
YOLOv8-X~\citep{Jocher_Ultralytics_YOLO_2023}      & 16.7 & 31.4 & 15.8 & 1.3 & 11.4 & 28.8 \\
YOLOv10-L~\citep{yolov10}     & 15.9 & 30.4 & 14.9 & 1.4 & 11.1 & 27.3 \\
YOLOv10-X~\citep{yolov10}     & 16.4 & 31.1 & 15.7 & 1.4 & 11.7 & 27.8 \\
YOLO11-L~\citep{yolo11_ultralytics}      & 16.7 & 31.6 & 16.0 & 1.3 & 11.4 & 28.8 \\
YOLO11-X~\citep{yolo11_ultralytics}      & 17.6 & 33.1 & 16.6 & 1.6 & 12.1 & 30.3 \\
YOLO12-L~\citep{tian2025yolov12}      & 16.6 & 31.3 & 15.7 & 1.2 & 11.5 & 28.8 \\
YOLO12-X~\citep{tian2025yolov12}      & 17.2 & 32.7 & 16.1 & 1.5 & 12.2 & 29.2 \\
\midrule
RT-DETR-R18~\citep{rtdetr}           & 16.9 & 31.5 & 15.3 & 2.9 & 13.0 & 26.2 \\
RT-DETR-R50~\citep{rtdetr}           & 17.8 & 34.5 & 16.1 & 2.9 & 13.6 & 28.5 \\
DEIM-RT-DETRv2-R18~\citep{huang2024deim}      & 16.7 & 32.1 & 15.3 & 3.2 & 13.5 & 27.3 \\
DEIM-RT-DETRv2-R50~\citep{huang2024deim}     & 17.2 & 33.6 & 15.7 & 2.6 & 13.3 & 28.5 \\
\midrule
\textbf{EFSI-DETR}    & \textbf{20.2} & \textbf{38.4} & \textbf{18.8} & \textbf{4.3} & \textbf{16.8} & \textbf{30.5} \\
\bottomrule
\end{tabular}

\end{table}


\paragraph{Comparison Results on CODrone}
Table~\ref{tab_codrone} presents the evaluation results on the CODrone dataset, where our EFSI-DETR achieves superior performance across key metrics. Compared to YOLO series methods, EFSI-DETR obtains 20.2\% AP, outperforming YOLOv12-X by 3.0\% and YOLO11-X\citep{Jocher_Ultralytics_YOLO_2023} by 2.6\%. The improvements are more evident in AP$_{50}$, where EFSI-DETR attains 38.4\%, outperforming YOLOv12-X and YOLO11-X, which achieve 32.7\% and 33.1\%, respectively.  For small object detection, EFSI-DETR demonstrates consistent improvements with 4.3\% AP$_s$, surpassing YOLOv12-X by 2.8\% and YOLO11-X by 2.7\%.

\begin{figure*} \centering
    \includegraphics[width=0.24\textwidth]{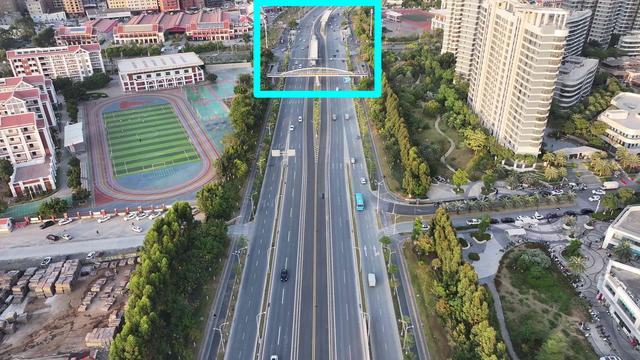}\hspace{-1mm}
    \includegraphics[width=0.24\textwidth]{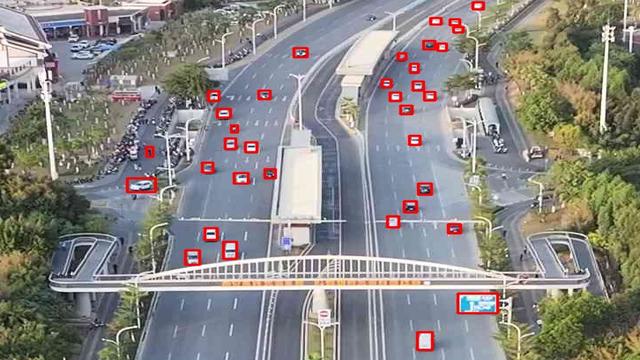}\hspace{-1mm}
    \includegraphics[width=0.24\textwidth]{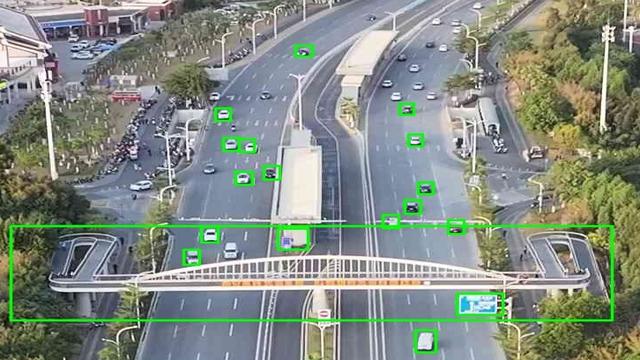}\hspace{-1mm}
    \includegraphics[width=0.24\textwidth]{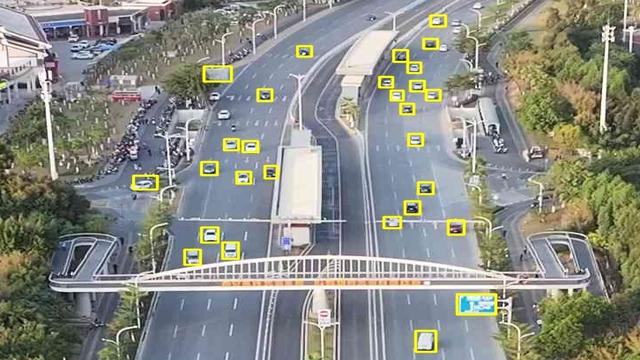}
    \\
    \includegraphics[width=0.24\textwidth]{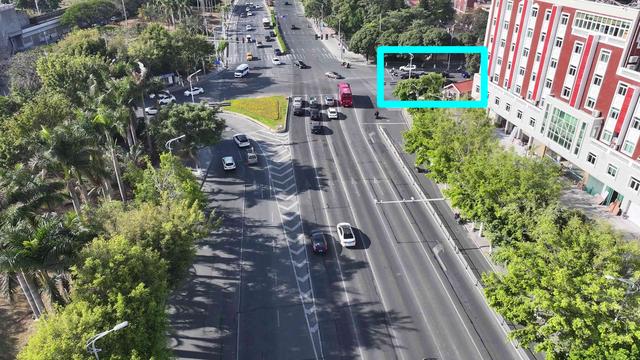}\hspace{-1mm}
    \includegraphics[width=0.24\textwidth]{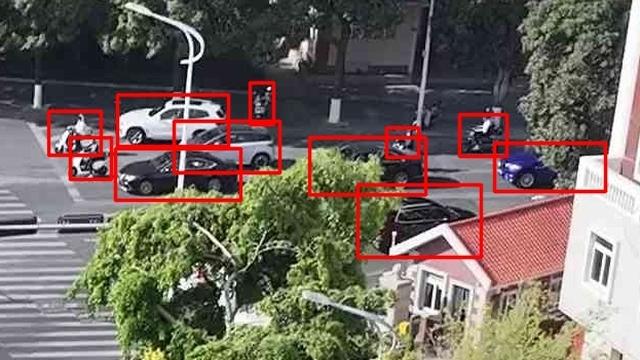}\hspace{-1mm}
    \includegraphics[width=0.24\textwidth]{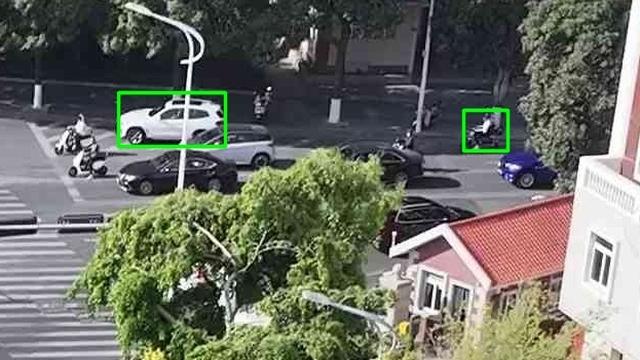}\hspace{-1mm}
    \includegraphics[width=0.24\textwidth]{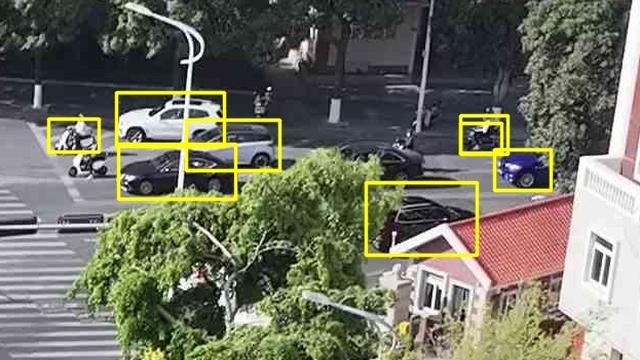}
    \\
    \includegraphics[width=0.24\textwidth]{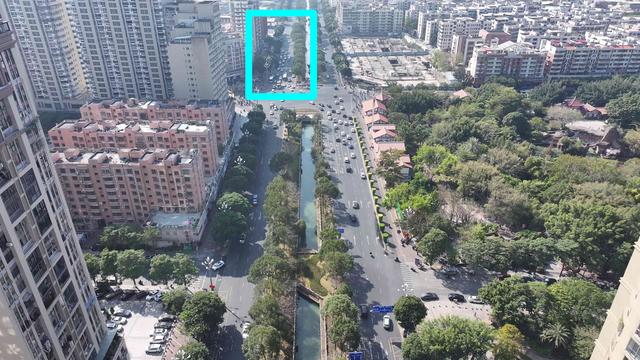}\hspace{-1mm}
    \includegraphics[width=0.24\textwidth]{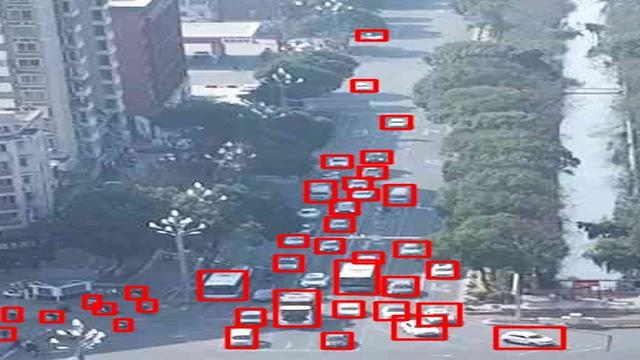}\hspace{-1mm}
    \includegraphics[width=0.24\textwidth]{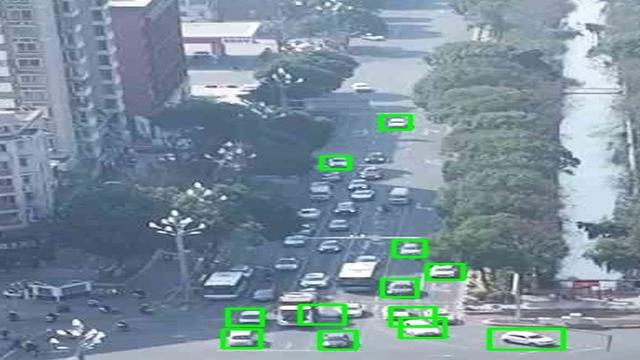}\hspace{-1mm}
    \includegraphics[width=0.24\textwidth]{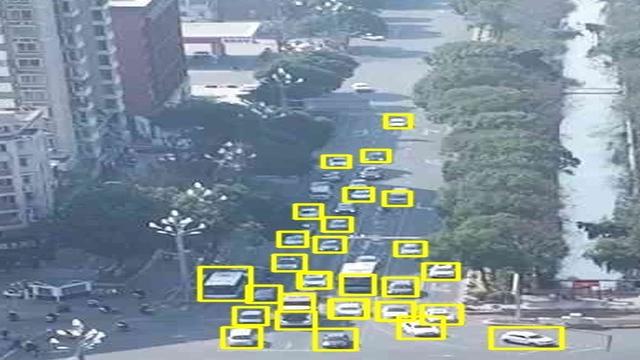}
    \\
        \includegraphics[width=0.24\textwidth]{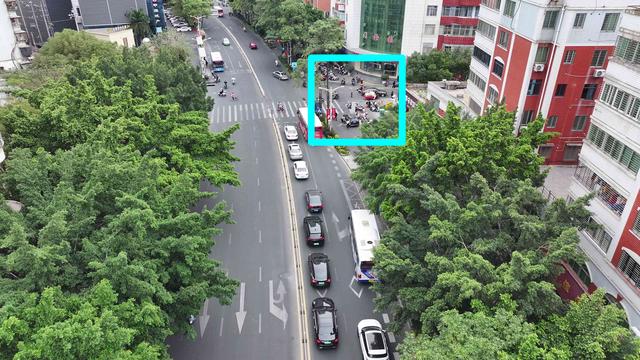}\hspace{-1mm}
    \includegraphics[width=0.24\textwidth]{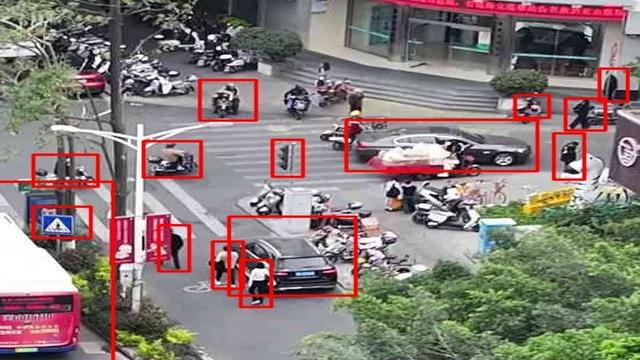}\hspace{-1mm}
    \includegraphics[width=0.24\textwidth]{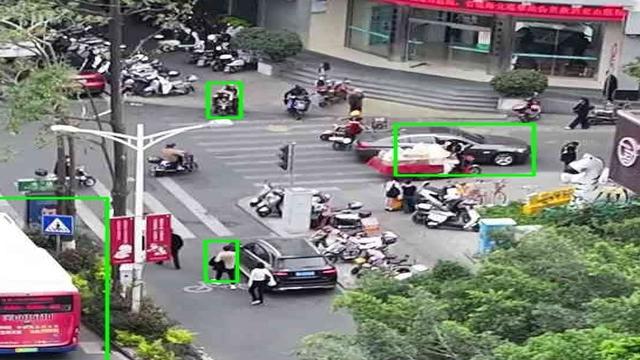}\hspace{-1mm}
    \includegraphics[width=0.24\textwidth]{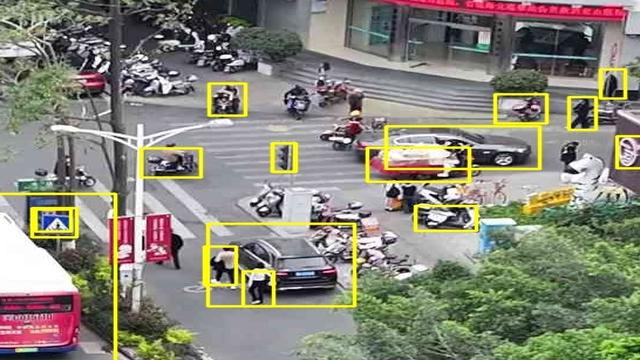}
    \\
    \includegraphics[width=0.24\textwidth]{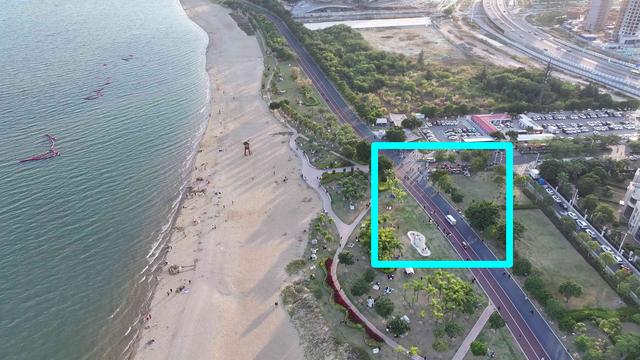}\hspace{-1mm}
    \includegraphics[width=0.24\textwidth]{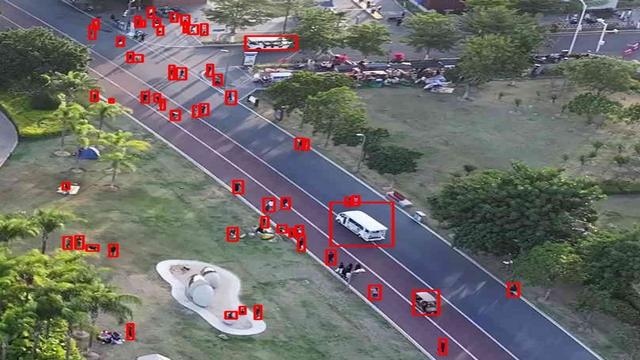}\hspace{-1mm}
    \includegraphics[width=0.24\textwidth]{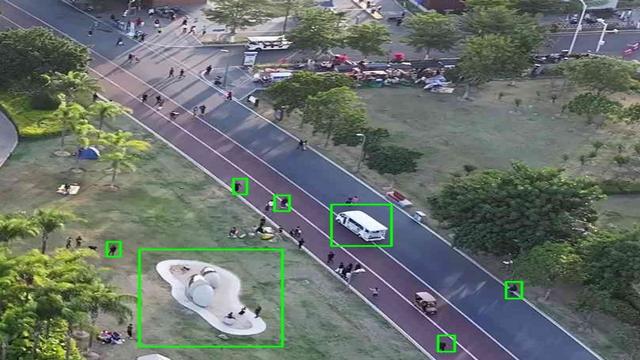}\hspace{-1mm}
    \includegraphics[width=0.24\textwidth]{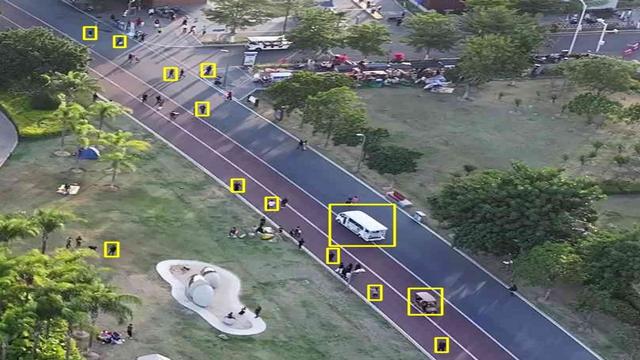}
    \\
    \includegraphics[width=0.24\textwidth]{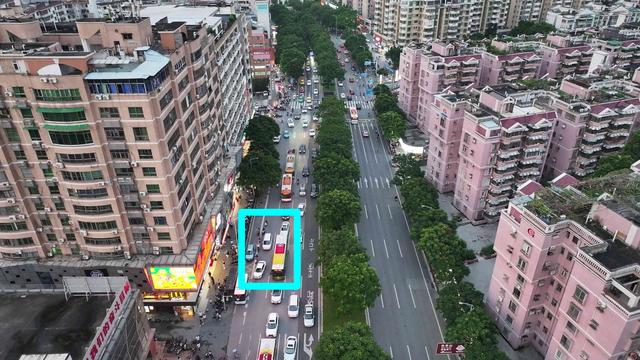}\hspace{-1mm}
    \includegraphics[width=0.24\textwidth]{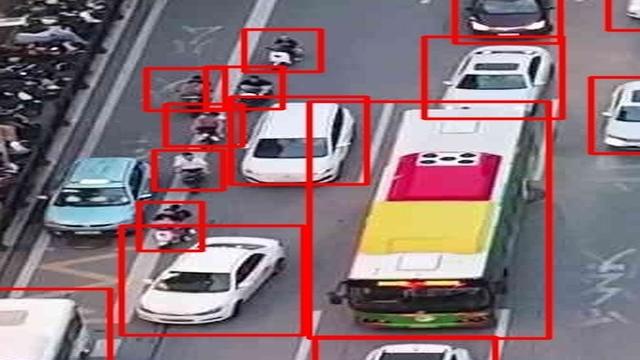}\hspace{-1mm}
    \includegraphics[width=0.24\textwidth]{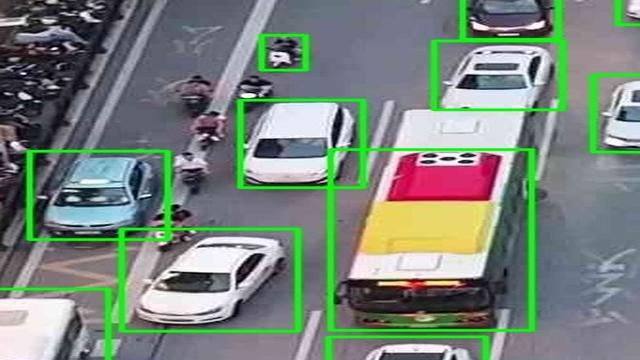}\hspace{-1mm}
    \includegraphics[width=0.24\textwidth]{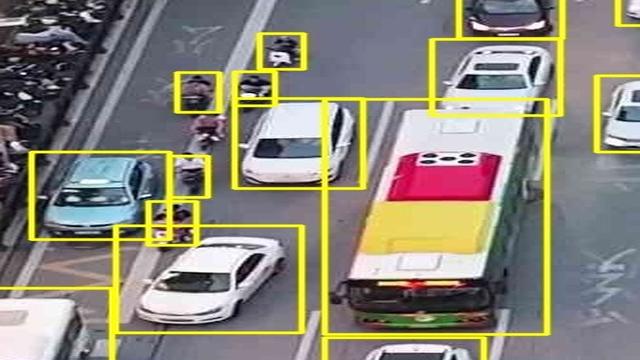}
    \makebox[0.24\textwidth]{\scriptsize \textbf{}}\hspace{-1mm}
    \makebox[0.24\textwidth]{\scriptsize \textbf{GT}}\hspace{-1mm}
    \makebox[0.24\textwidth]{\scriptsize \textbf{RT-DETR}}\hspace{-1mm}
    \makebox[0.24\textwidth]{\scriptsize \textbf{EFSI-DETR (Ours)}}
    \\
    \caption{Visualization of detection results of RT-DETR and EFSI-DETR.} 
    \label{fig_vis_result}
\end{figure*}

When compared to DETR-based approaches, EFSI-DETR exhibits equally noteworthy benefits. Our method exceeds RT-DETR-R50 by 2.4\% in AP and 3.9\% in AP$_{50}$, while outperforming DEIM-RT-DETRv2-R50 by 3.0\% in AP and 4.8\% in AP$_{50}$. The improvements in small object detection are particularly notable, with EFSI-DETR achieving 1.4\% and 1.7\% gains in AP$_s$ over RT-DETR-R50 and DEIM-RT-DETRv2-R50 respectively. A comparative visualization of detection results is presented in Fig.~\ref{fig_vis_result}. These results further substantiate the effectiveness of the proposed method, demonstrating its robust generalization capability across a wide range of drone-based object detection scenarios.

\begin{table}[ht]
\centering
\small 
\setlength{\tabcolsep}{1.2pt} 
\caption{Experiments on VisDrone dataset to evaluate the individual contributions of the key components. The variants are denoted as \textbf{V$_A$}, \textbf{V$_B$}, \textbf{V$_C$}, and \textbf{V$_D$}.}
\label{tab_ablation}
\begin{tabular}{c|ccc|ccc|c}
\toprule
\textbf{Method} & \textbf{FFR} & \textbf{DyFusNet} & \textbf{ESFC} & \textbf{AP$^{val}$} & \textbf{AP$^{val}_{50}$} & \textbf{AP$^{val}_s$} & \textbf{Params(M)} \\
\midrule
\textbf{V$_A$} & \ding{55} & \ding{55} & \ding{55} & 26.9 & 44.5 & 18.3 & 25.6 \\
\textbf{V$_B$} & \ding{51} & \ding{55} & \ding{55} & 31.3 & 50.6 & 23.2 & 27.7\\
\textbf{V$_C$} & \ding{51} & \ding{51} & \ding{55} & 32.7 & 52.3 & 24.6 & 28.8\\
\textbf{V$_D$} & \ding{51} & \ding{51} & \ding{51} & \textbf{33.1} & \textbf{52.7} & \textbf{24.8} & 27.3 \\
\bottomrule
\end{tabular}
\end{table}

\subsection{Visualization of Per-category Precision}
\begin{figure*} \centering
    \includegraphics[width=0.45\textwidth]{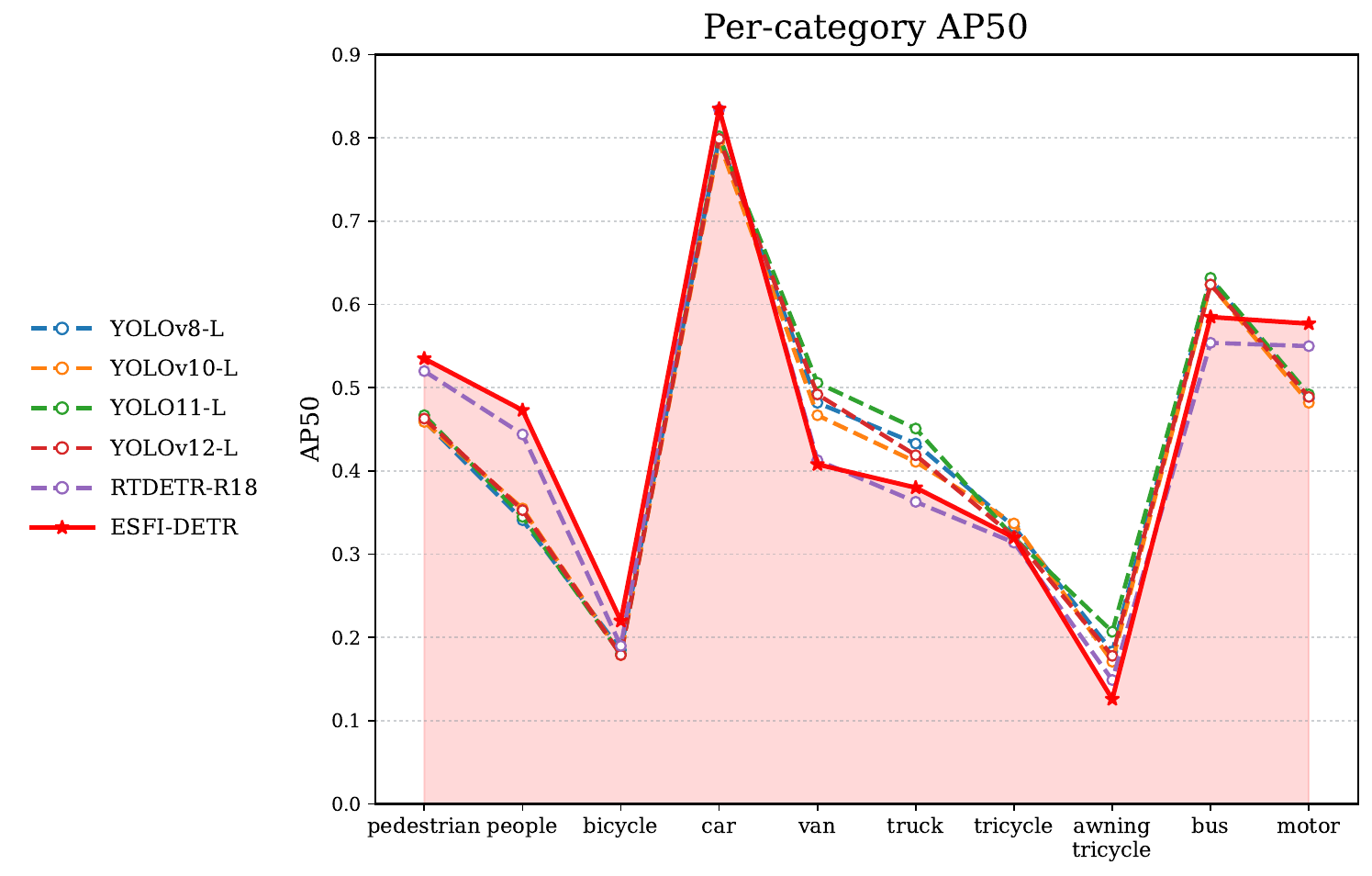}\hspace{-1mm}
    \includegraphics[width=0.45\textwidth]{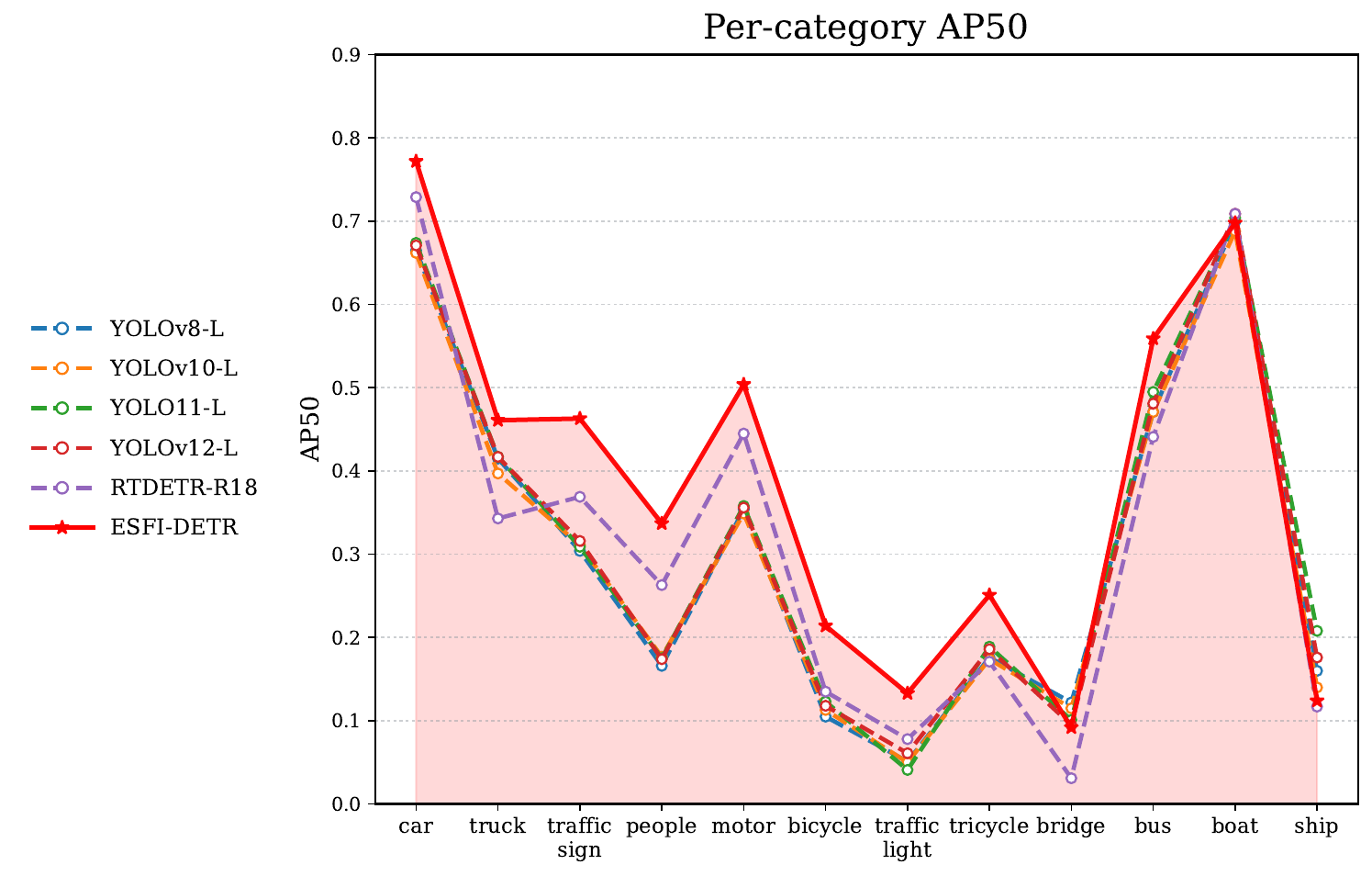}
    \\
    \includegraphics[width=0.45\textwidth]{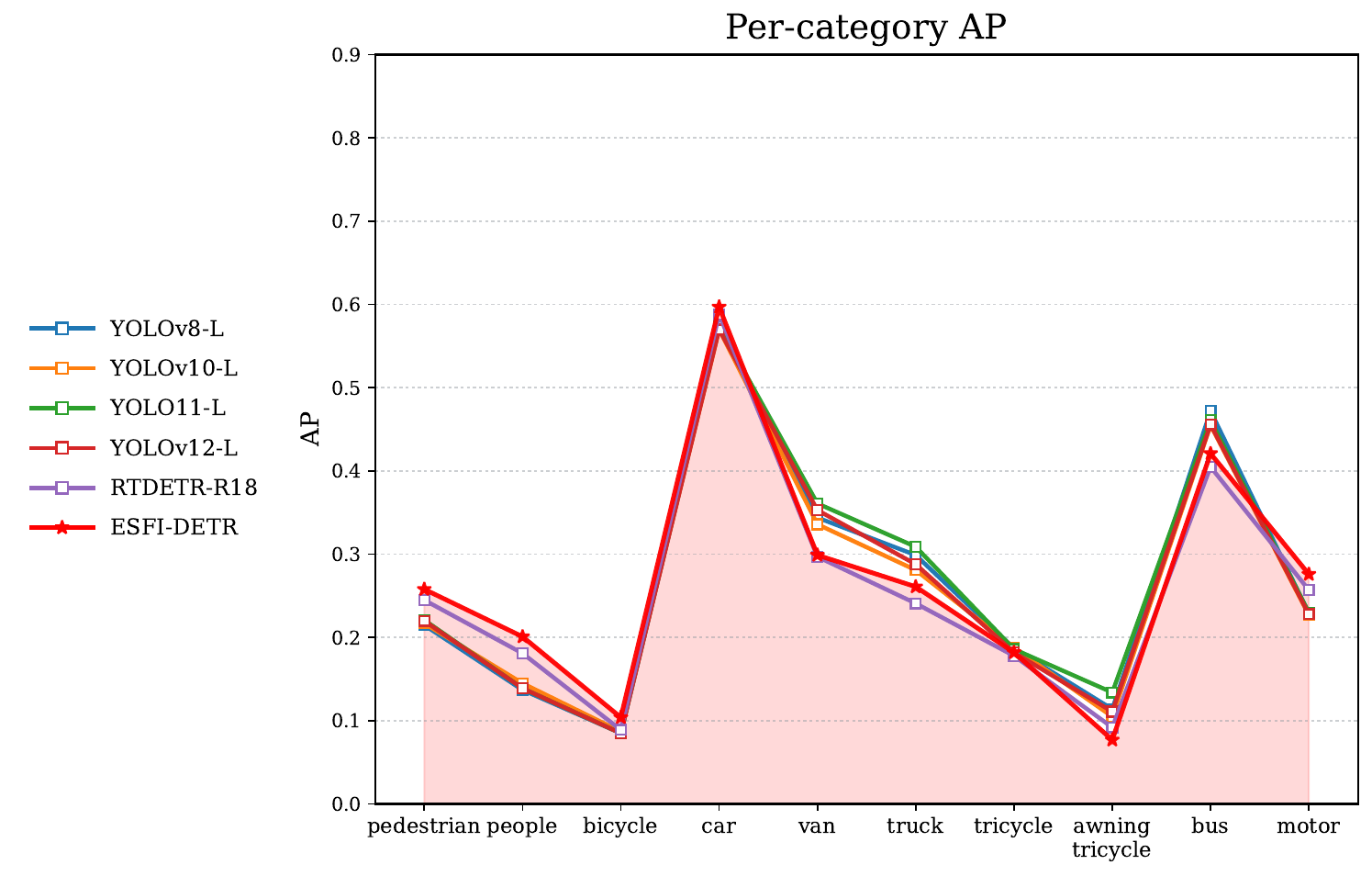}\hspace{-1mm}
    \includegraphics[width=0.45\textwidth]{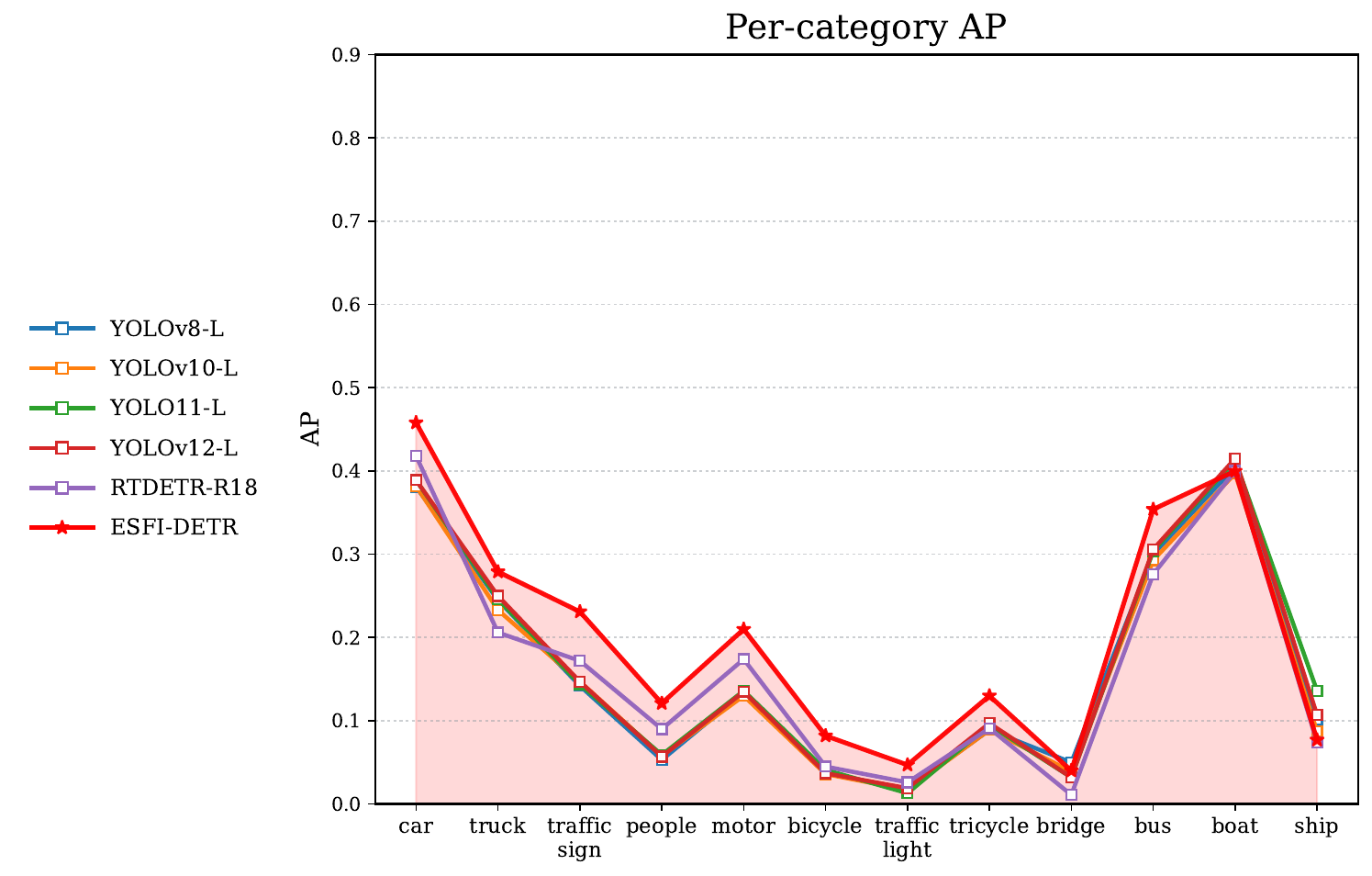}
    \\
    \includegraphics[width=0.45\textwidth]{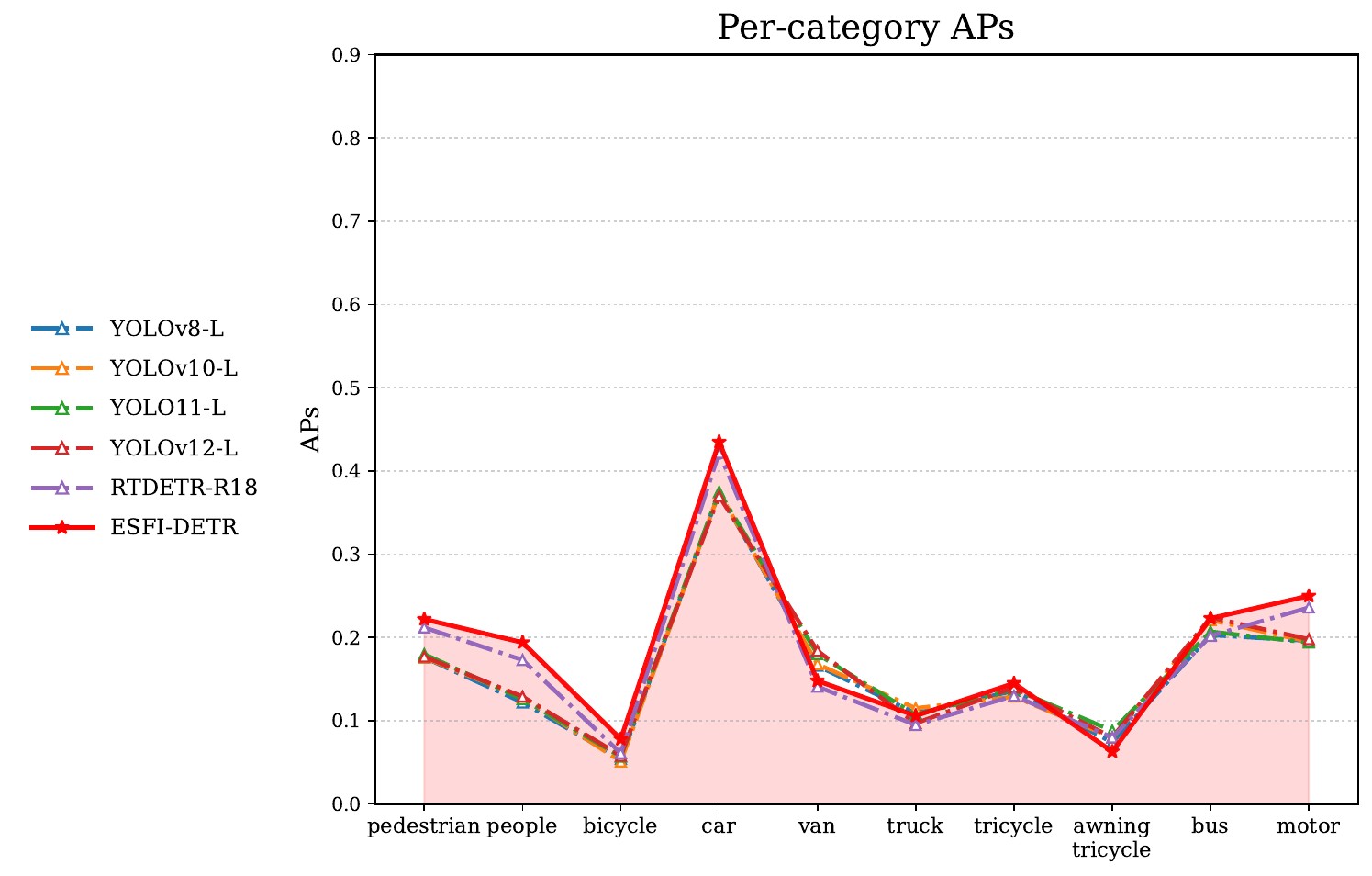}\hspace{-1mm}
    \includegraphics[width=0.45\textwidth]{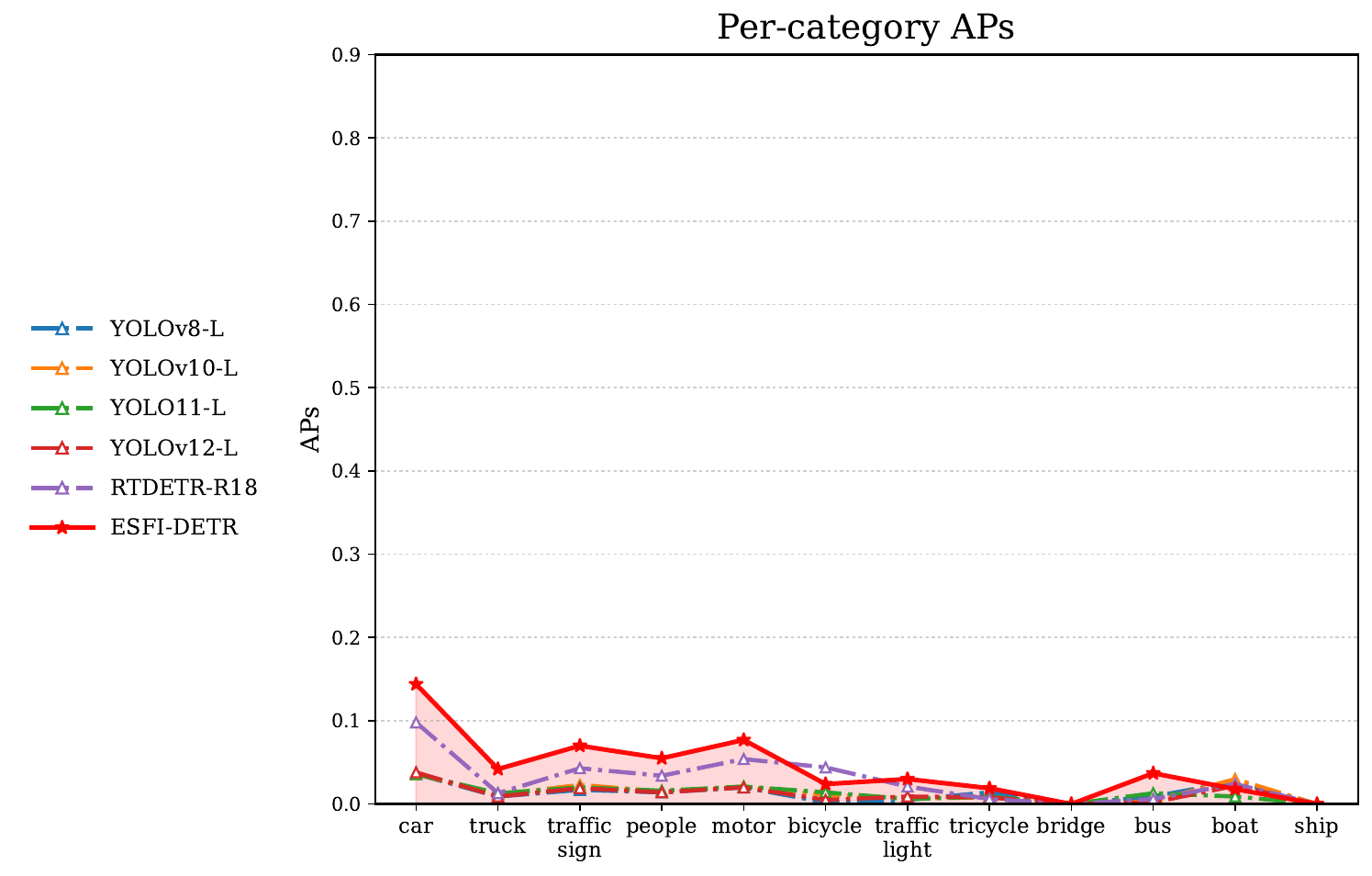}
    \\
    \makebox[0.45\textwidth]{\textbf{VisDrone}}\hspace{-1mm}
    \makebox[0.45\textwidth]{\textbf{CODrone}}
    \caption{Line chart visualization of AP, AP$_{50}$ and AP$_s$ across categories on VisDrone and CODrone datasets.} 
    \label{fig_vis_line}
\end{figure*}

Fig.~\ref{fig_vis_line} provides a category-level comparison of AP, AP$_{50}$, and AP$_s$ on VisDrone and CODrone. Overall, EFSI-DETR’s curves are more frequently located at or near the upper range across categories, with fewer sharp degradations than several baselines. On VisDrone, the relative gains are more apparent on small or clutter-prone categories (e.g., pedestrian/people/bicycle/motor), while the performance on dominant vehicle classes (e.g., car/van/truck/bus) remains comparable. On CODrone, a similar pattern is observed: EFSI-DETR shows clearer improvements on thin-structure or low-pixel-support categories (e.g., traffic sign/light, bicycle) and maintains competitive results on the common large classes (e.g., car, bus, truck). The advantage is most evident in AP$_s$, indicating that better small-object performance achieved at the category level, whereas AP/AP$_{50}$ gains are more moderate and class-dependent.

\subsection{Per-class Precision-Recall Analysis}
\begin{figure*}[!htbp] \centering
    \includegraphics[width=0.49\textwidth]{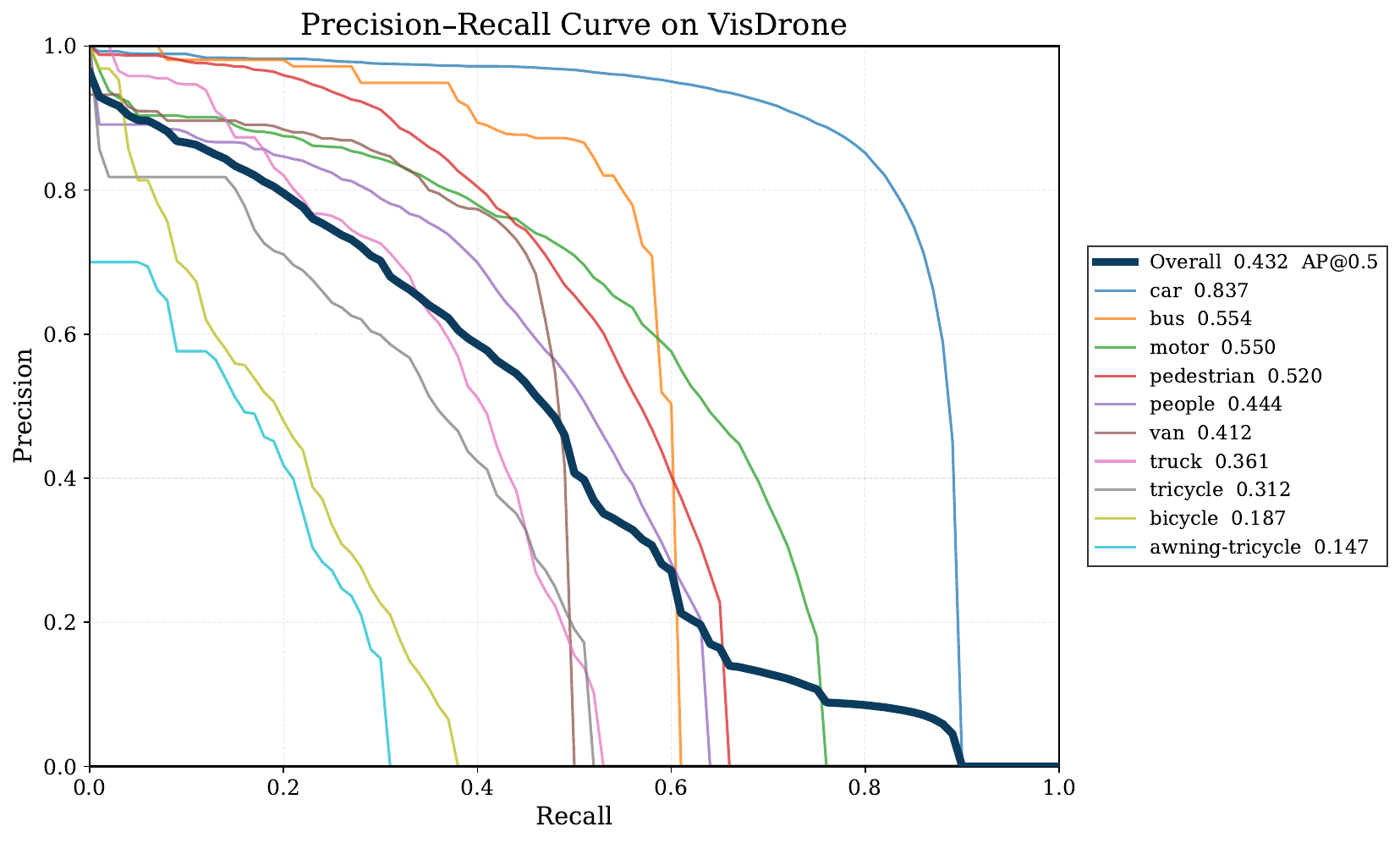}\hspace{-1mm}
    \includegraphics[width=0.49\textwidth]{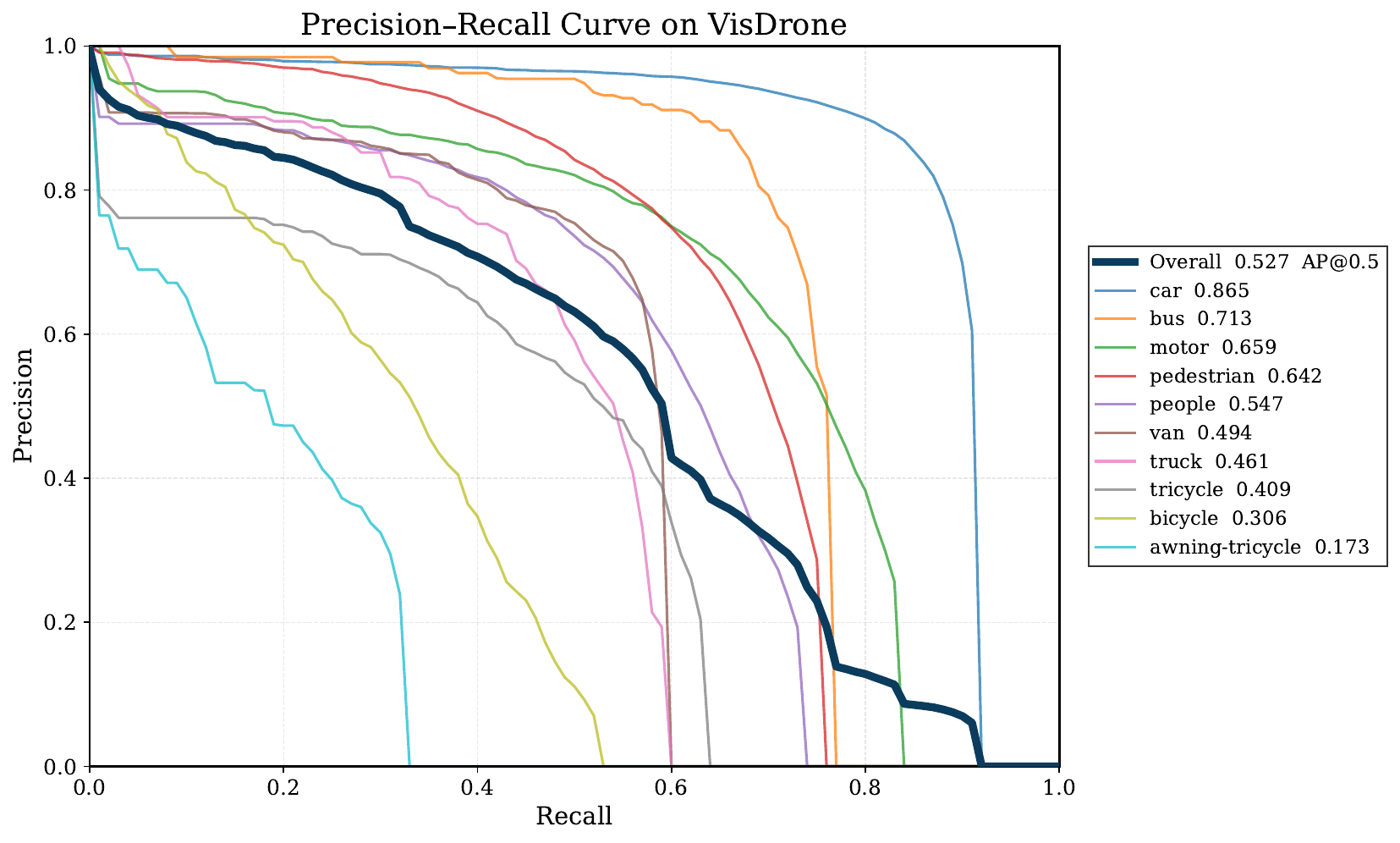}
    \\
    \includegraphics[width=0.49\textwidth]{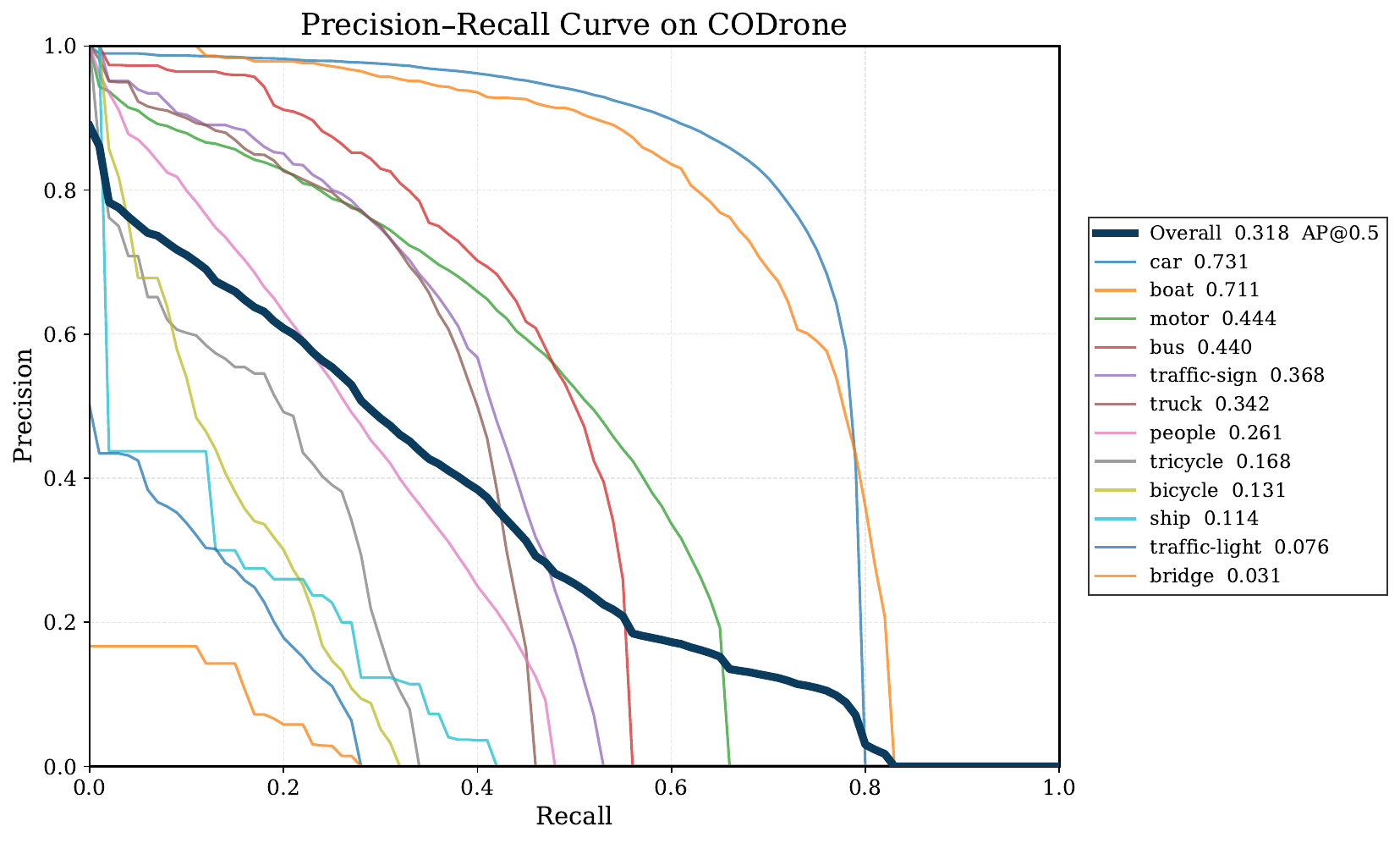}\hspace{-1mm}
    \includegraphics[width=0.49\textwidth]{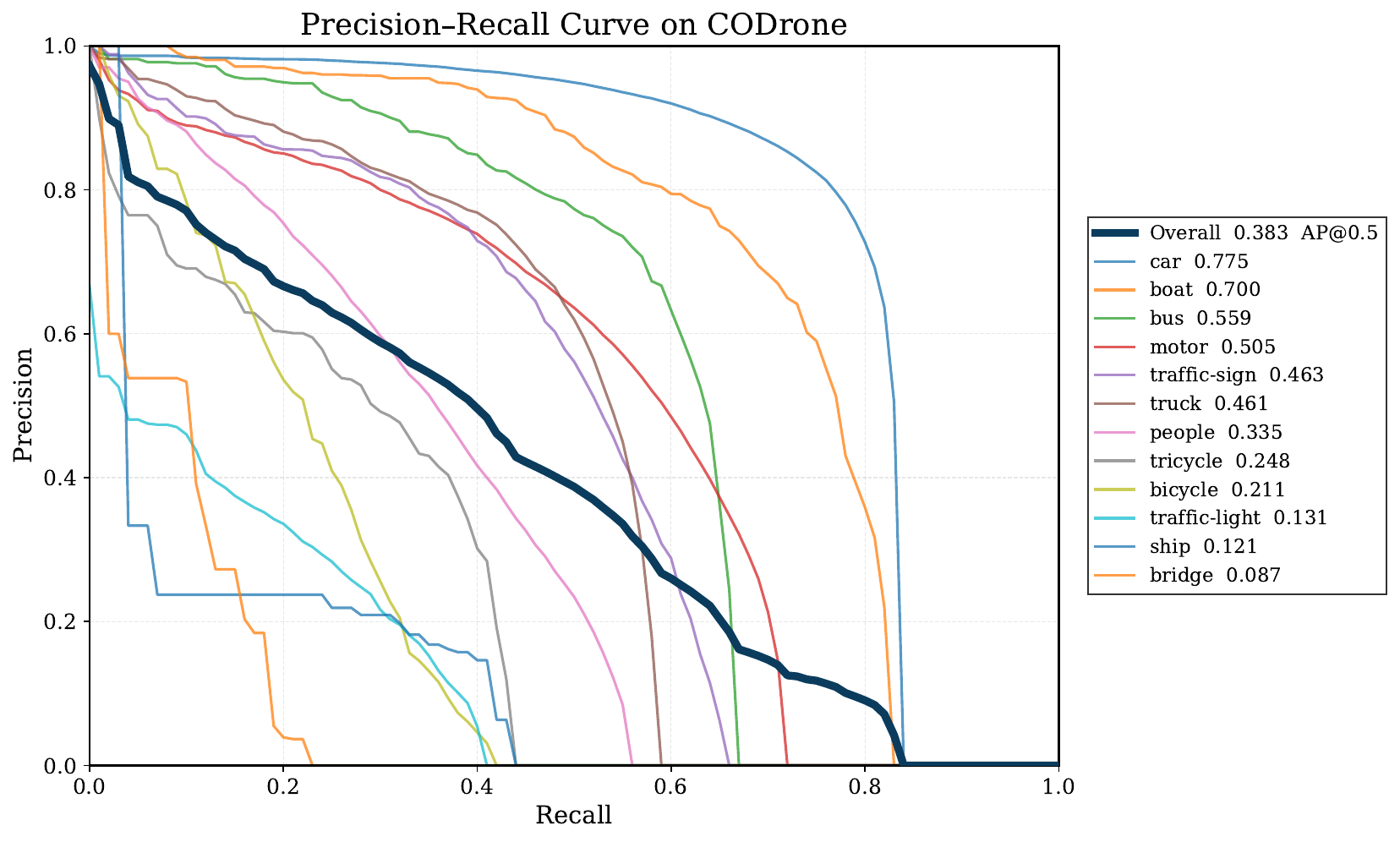}

    \makebox[0.49\textwidth]{\scriptsize \textbf{RT-DETR}}\hspace{-1mm}
    \makebox[0.49\textwidth]{\scriptsize \textbf{EFSI-DETR (Ours)}}
    \\
    \caption{Per-class Precision-Recall (PR) Curves of RT-DETR and EFSI-DETR on the VisDrone and CODrone.} 
    \label{fig_vis_pr}
\end{figure*}

Fig.\ref{fig_vis_pr} illustrates the class-wise precision-recall (PR) curves for RT-DETR and EFSI-DETR on the VisDrone and CODrone datasets. It is observed that relatively large-object categories (e.g., car) achieve near-optimal performance in the upper-right corner of the PR space—where precision and recall reach their respective maxima—thus leaving a negligible margin for further performance improvement. In contrast, categories dominated by small objects and susceptible to background clutter (e.g., bicycle, tricycle, pedestrian, traffic light, bridge) exhibit premature precision degradation and suffer from severe recall bottlenecks for the baseline RT-DETR. EFSI-DETR consistently elevates the PR envelopes of these challenging categories within the mid-to-high recall regime and delivers significant AP$_{50}$ performance gains on such categories. This finding corroborates that the overall performance gains of EFSI-DETR originate predominantly from enhanced recall on hard instances while maintaining comparable precision levels, rather than a uniform accuracy boost for saturated, high-performance categories.

\subsection{Failure Case Analysis}
\begin{figure*} \centering
    \includegraphics[width=0.24\textwidth]{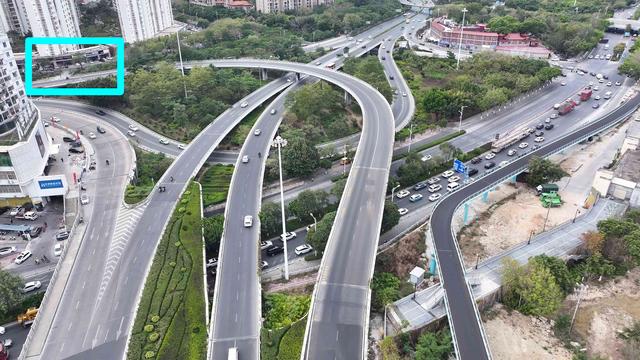}\hspace{-1mm}
    \includegraphics[width=0.24\textwidth]{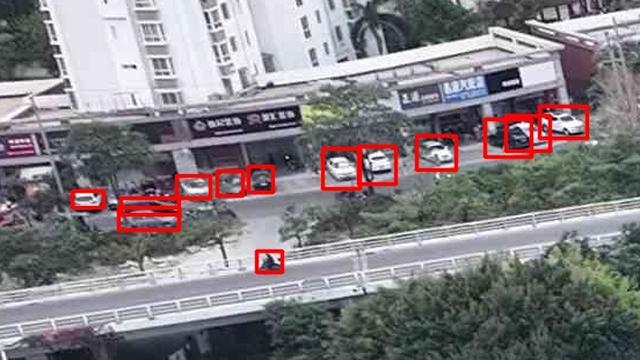}\hspace{-1mm}
    \includegraphics[width=0.24\textwidth]{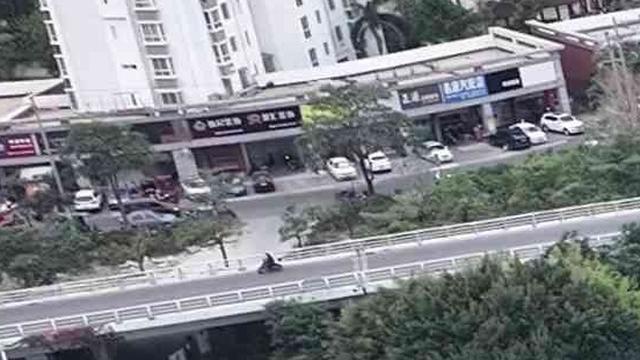}\hspace{-1mm}
    \includegraphics[width=0.24\textwidth]{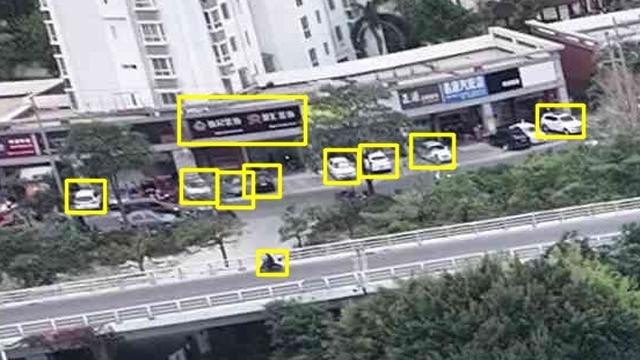}
    \\

    \includegraphics[width=0.24\textwidth]{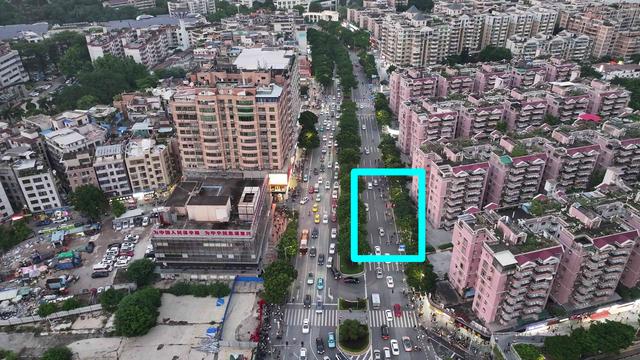}\hspace{-1mm}
    \includegraphics[width=0.24\textwidth]{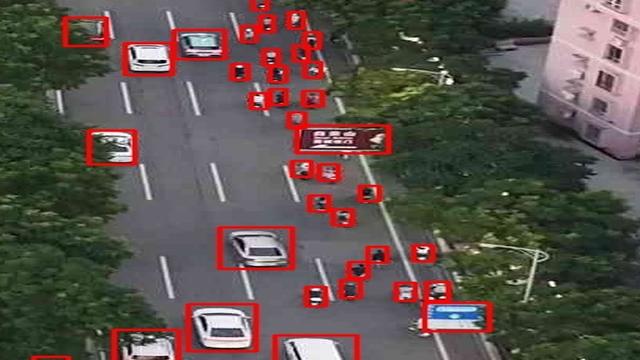}\hspace{-1mm}
    \includegraphics[width=0.24\textwidth]{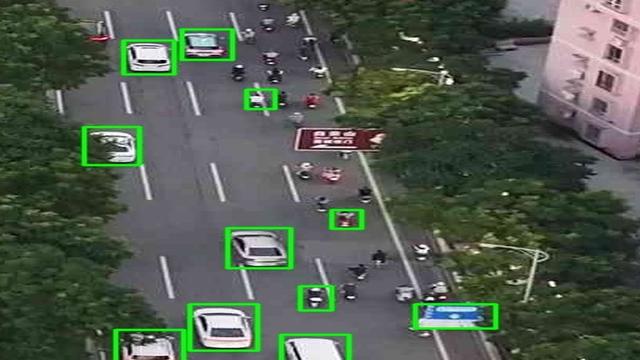}\hspace{-1mm}
    \includegraphics[width=0.24\textwidth]{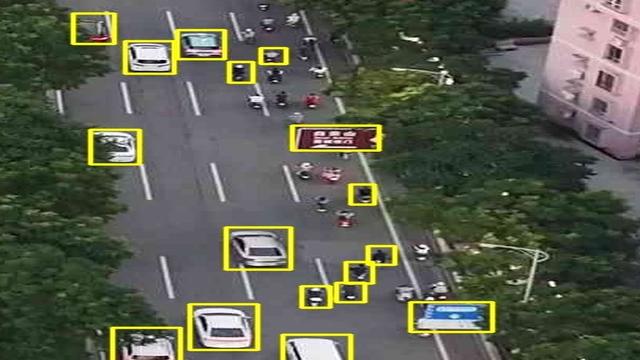}
    \\
    \includegraphics[width=0.24\textwidth]{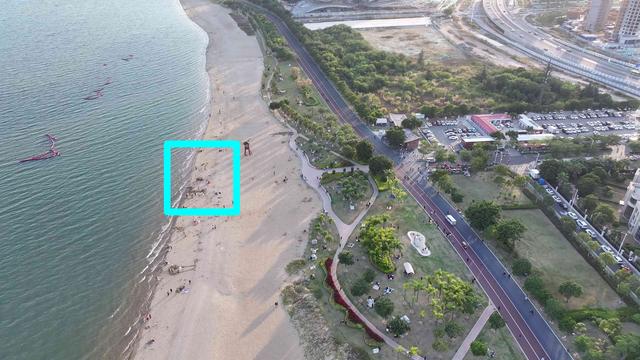}\hspace{-1mm}
    \includegraphics[width=0.24\textwidth]{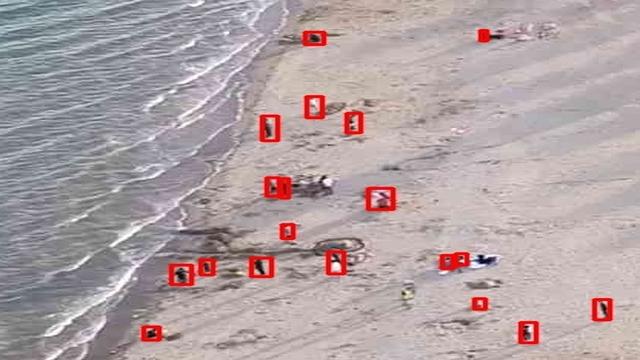}\hspace{-1mm}
    \includegraphics[width=0.24\textwidth]{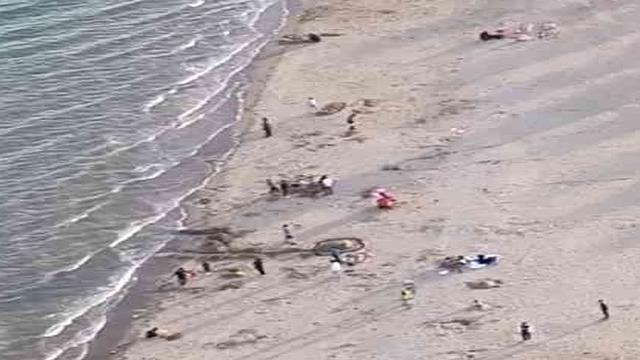}\hspace{-1mm}
    \includegraphics[width=0.24\textwidth]{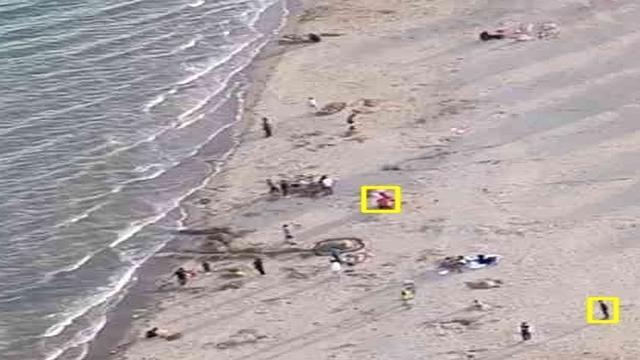}

    \makebox[0.24\textwidth]{\scriptsize \textbf{}}\hspace{-1mm}
    \makebox[0.24\textwidth]{\scriptsize \textbf{GT}}\hspace{-1mm}
    \makebox[0.24\textwidth]{\scriptsize \textbf{RT-DETR}}\hspace{-1mm}
    \makebox[0.24\textwidth]{\scriptsize \textbf{EFSI-DETR (Ours)}}
    \\
    \caption{Failure Case Visualization of RT-DETR and EFSI-DETR.} 
    \label{fig_vis_badcase}
\end{figure*}

While EFSI-DETR achieves consistent performance gains on mainstream UAV detection benchmarks, and although our method has already achieved notably better performance than RT-DETR, it still generates suboptimal predictions in several challenging scenarios, as visualized in Fig. \ref{fig_vis_badcase}. First, extremely tiny instances (e.g., targets composed of merely a few pixels with extremely low contrast against the background) remain a major challenge: although the FFR module explicitly retains high-resolution shallow features (\textbf{S}$_1$/\textbf{S}$_2$) and prioritizes fine-grained multi-scale features (\textbf{F}$_2$-\textbf{F}$_4$) for precise localization, the inherent scarcity of discriminative visual evidence in such scenarios often leads to missed detections or unstable localization results.
Second, in scenarios with heavy occlusion or extreme crowding, multiple target instances overlap heavily within a small spatial area. This introduces significant assignment ambiguity in the query-based matching mechanism of our detector, which may further result in false negatives or duplicate predictions.  

\subsection{Ablation Study}
To validate the effectiveness of core module design in EFSI-DETR, we design a series of ablation experiments on the VisDrone dataset. We use RT-DETR-R18$_{\epsilon=1}$ as a baseline model in all ablation experiments, where $\epsilon$ represents the expression ratio of the channel in the fusion stage.

\paragraph{Effect of key components}

\begin{figure}[!htbp] \centering
    \makebox[0.095\textwidth]{\scriptsize \textbf{GT}}\hspace{-1mm}
    \makebox[0.095\textwidth]{\scriptsize \textbf{V$_A$}}\hspace{-1mm}
    \makebox[0.095\textwidth]{\scriptsize \textbf{V$_B$}}\hspace{-1mm}
    \makebox[0.095\textwidth]{\scriptsize \textbf{V$_C$}}\hspace{-1mm}
    \makebox[0.095\textwidth]{\scriptsize \textbf{V$_D$}}
    \\
    \includegraphics[width=0.095\textwidth]{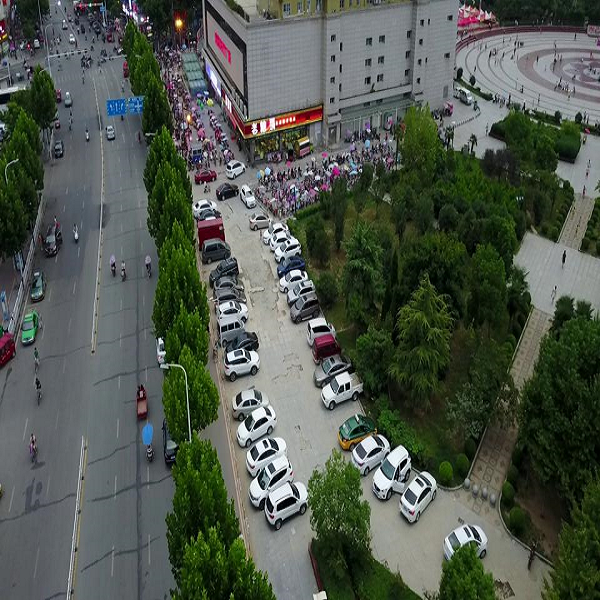}\hspace{-1mm}
    \includegraphics[width=0.095\textwidth]{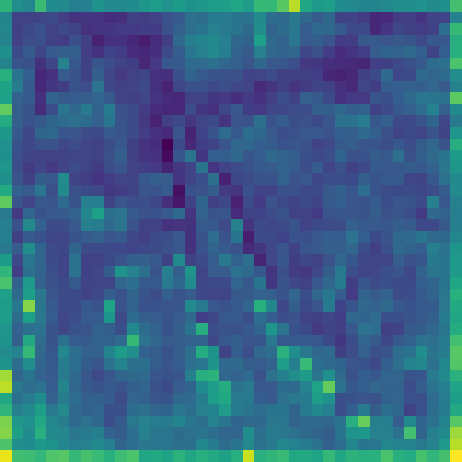}\hspace{-1mm}
    \includegraphics[width=0.095\textwidth]{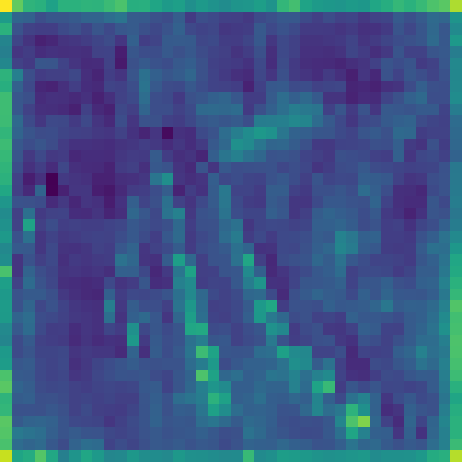}\hspace{-1mm}
    \includegraphics[width=0.095\textwidth]{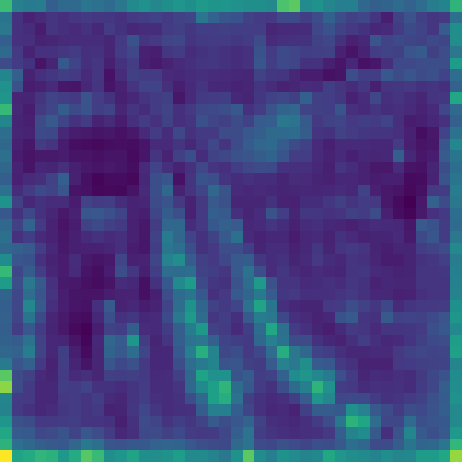}\hspace{-1mm}
    \includegraphics[width=0.095\textwidth]{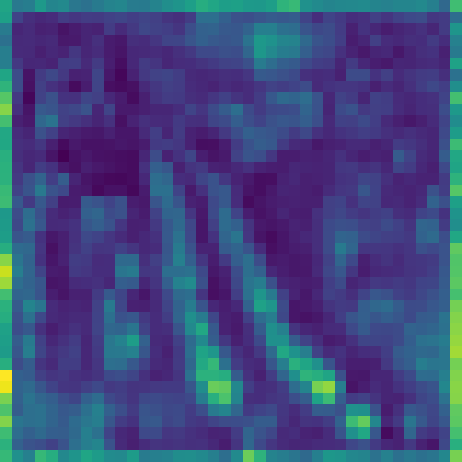}
    \\
    \includegraphics[width=0.095\textwidth]{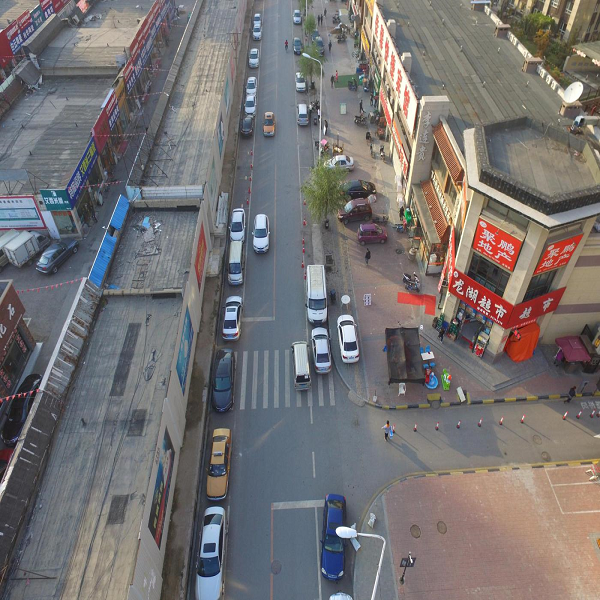}\hspace{-1mm}
    \includegraphics[width=0.095\textwidth]{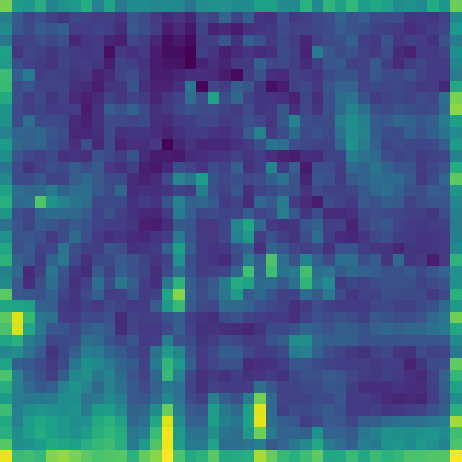}\hspace{-1mm}
    \includegraphics[width=0.095\textwidth]{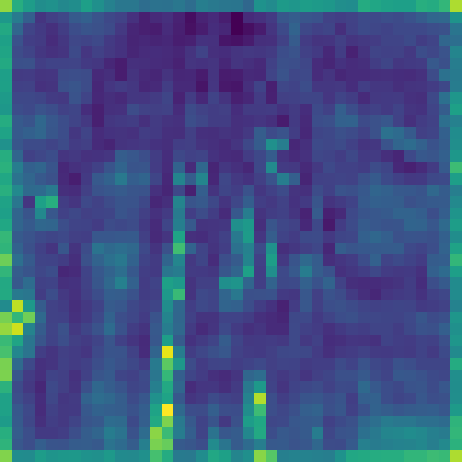}\hspace{-1mm}
    \includegraphics[width=0.095\textwidth]{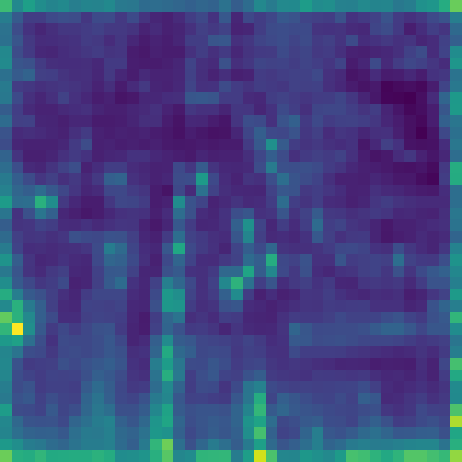}\hspace{-1mm}
    \includegraphics[width=0.095\textwidth]{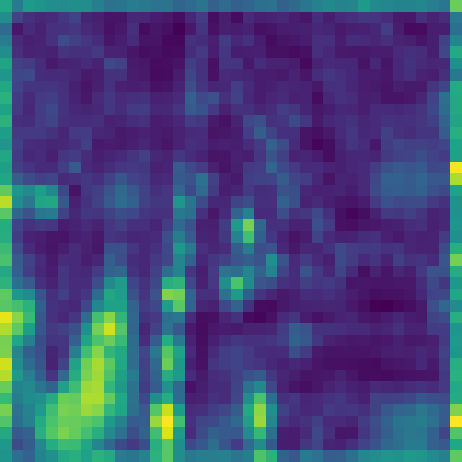}
    \\
    \includegraphics[width=0.095\textwidth]{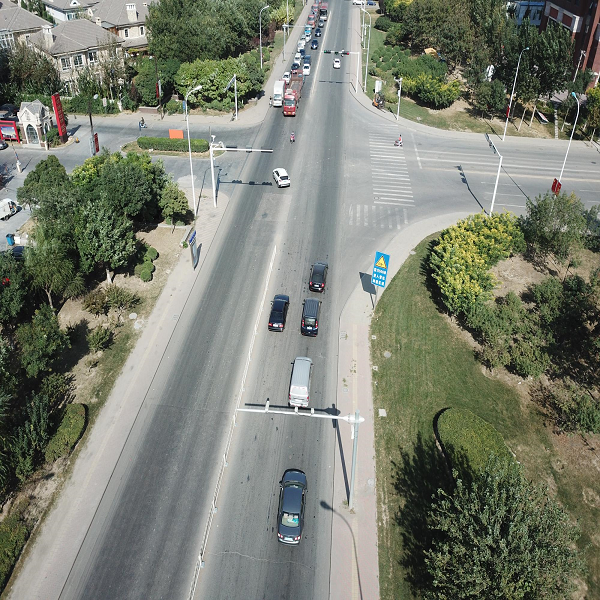}\hspace{-1mm}
    \includegraphics[width=0.095\textwidth]{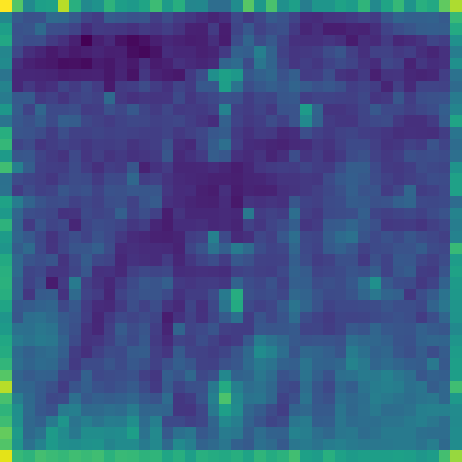}\hspace{-1mm}
    \includegraphics[width=0.095\textwidth]{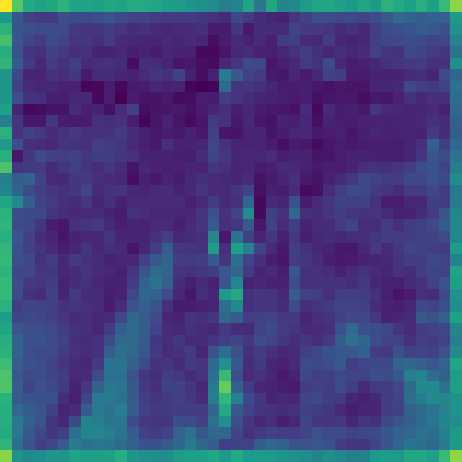}\hspace{-1mm}
    \includegraphics[width=0.095\textwidth]{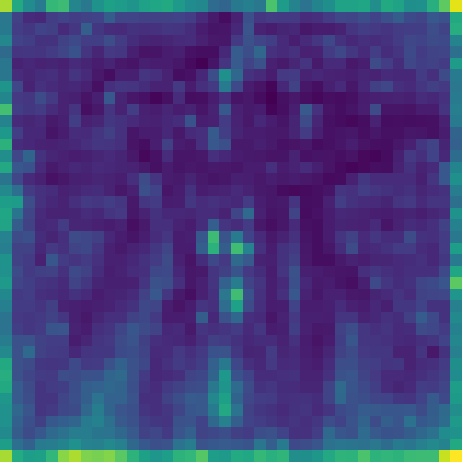}\hspace{-1mm}
    \includegraphics[width=0.095\textwidth]{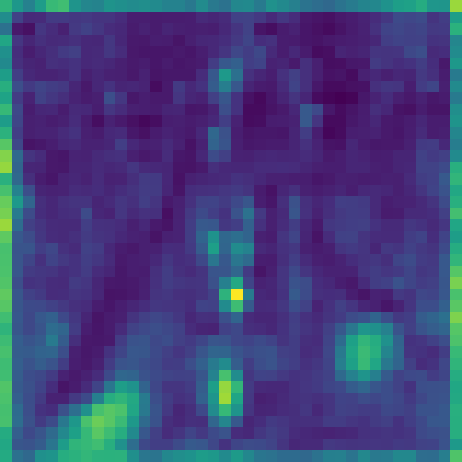}
    \\
    \caption{Visualization of feature maps for variants composed of different components. The results demonstrate the progressive improvement in feature representation and object awareness with the integration of proposed components.} 
    \label{fig_vis_fea}
\end{figure}

Experimental results presented in Table~\ref{tab_ablation} validate the effectiveness of each proposed component. Specifically, the introduction of the FFR strategy significantly enhances spatial detail preservation, leading to improvements of 4.4\%, 6.1\%, and 4.9\% in AP, AP$_{50}$, and AP$_s$, respectively. The incorporation of DyFusNet, which integrates frequency-aware decomposition with spatial multi-scale processing, further improves feature representation, resulting in additional gains of 1.4\% in AP, 1.7\% in AP$_{50}$, and 1.4\% in AP$_s$. Furthermore, the ESFC module enhances the semantic quality of features, delivering incremental improvements of 0.4\%, 0.4\%, and 0.2\% in AP, AP$_{50}$, and AP$_s$, while simultaneously reducing the parameter count by 1.5M. Collectively, these results demonstrate the effectiveness and efficiency of the proposed components, which significantly enhance detection performance while maintaining real-time capability. These findings are further supported by the qualitative comparisons shown in Fig.~\ref{fig_vis_fea}.

\paragraph{Comparison with transform-based alternatives}

To further validate the design choice of adopting a frequency-inspired yet non-transform formulation, we compare DyFusNet with two transform-based alternatives, namely a DWT-based variant and an FFT-based variant, while keeping the remaining architecture unchanged. As shown in Table~\ref{tab_dyfus_freq}, the original DyFusNet achieves the best overall AP and AP$_{50}$, outperforming the DWT-based variant by 1.0\% and 1.0\%, respectively, and the FFT-based variant by 0.8\% and 0.9\%.

\begin{figure}[!htbp]
\centering
\makebox[0.12\textwidth]{\scriptsize \textbf{GT}}\hspace{-1mm}
\makebox[0.12\textwidth]{\scriptsize \textbf{FFT}}\hspace{-1mm}
\makebox[0.12\textwidth]{\scriptsize \textbf{DWT}}\hspace{-1mm}
\makebox[0.12\textwidth]{\scriptsize \textbf{DyFusNet}}
\\
\includegraphics[width=0.12\textwidth]{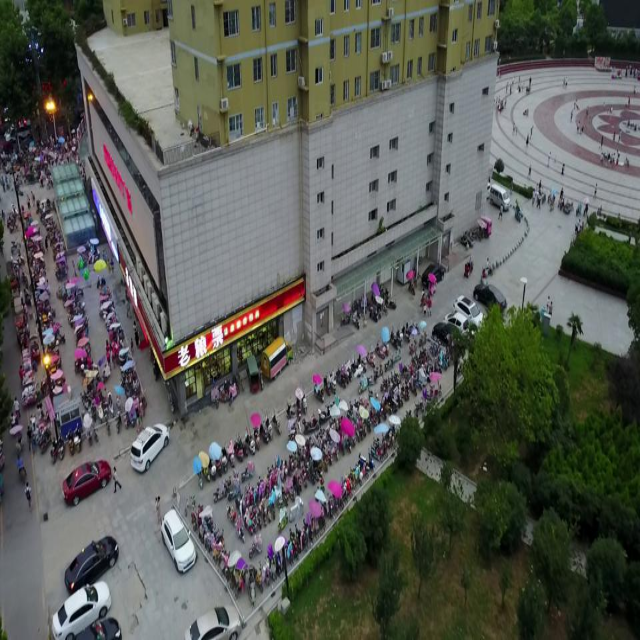}\hspace{-1mm}
\includegraphics[width=0.12\textwidth]{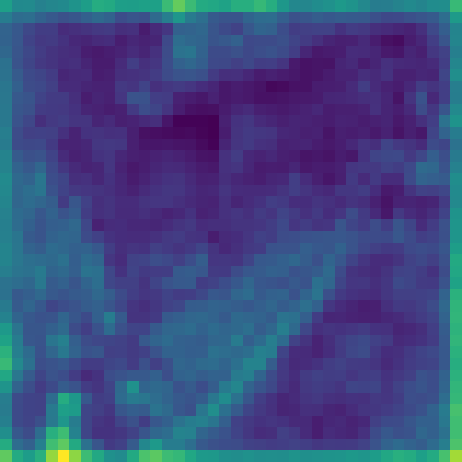}\hspace{-1mm}
\includegraphics[width=0.12\textwidth]{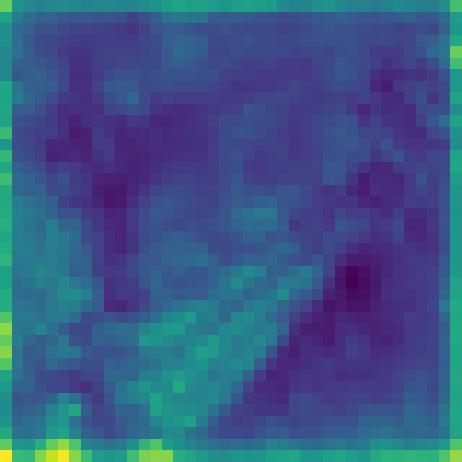}\hspace{-1mm}
\includegraphics[width=0.12\textwidth]{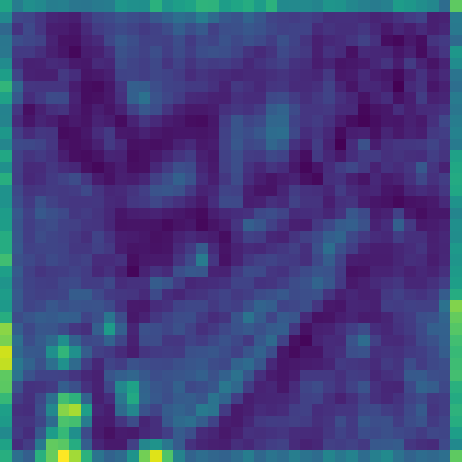}
\caption{Visualization of feature maps for methods with DyFusNet and transform-based alternatives. }
\label{fig_dyfus_freq_vis}
\end{figure}

These results are consistent with the motivation of DyFusNet discussed in Sec.~III-A. Instead of relying on explicit transform-domain processing, DyFusNet performs content-adaptive band emphasis directly in the spatial domain, thereby avoiding transform-related overhead while preserving discriminative frequency-selective behavior. As illustrated in Fig.~\ref{fig_dyfus_freq_vis}, DyFusNet yields more compact and object-focused responses in crowded UAV scenes, whereas the DWT- and FFT-based alternatives exhibit relatively more diffuse activations.

\begin{table}[ht]
\centering
\small
\setlength{\tabcolsep}{5pt}
\caption{Comparison between DyFusNet and transform-based alternatives.}
\label{tab_dyfus_freq}
\begin{tabular}{lccc}
\toprule
\textbf{Method} & \textbf{AP$^{val}$} & \textbf{AP$^{val}_{50}$} & \textbf{AP$^{val}_{s}$} \\
\midrule
EFSI-DETR (DyFusNet) & \textbf{33.1} & \textbf{52.7} & \textbf{24.8} \\
EFSI-DETR w/ DWT     & 32.1 & 51.7 & 24.4 \\
EFSI-DETR w/ FFT     & 32.3 & 51.8 & 24.8 \\
\bottomrule
\end{tabular}
\end{table}

\begin{table}[ht]
\centering
\small 
\setlength{\tabcolsep}{5pt} 
\caption{Results with different expert numbers in ESFC.}
\label{tab_esfc1}
\begin{tabular}{c|ccc|cc}
\toprule
\textbf{Experts} & \textbf{AP$^{val}$} & \textbf{AP$^{val}_{50}$} & \textbf{AP$^{val}_s$} & \textbf{Params(M)} & \textbf{GFLOPs} \\
\midrule
2 & 32.6 & 52.1 & 24.8 & 27.2 & 291.1 \\
3 & \textbf{33.1} & \textbf{52.7} & \textbf{24.8} & 27.3 & 291.5 \\
4 & 32.3 & 52.1 & 24.5 & 27.5 & 291.9 \\
5 & 32.6 & 51.9 & 24.4 & 27.6 & 292.3 \\
\bottomrule
\end{tabular}

\end{table}

\paragraph{Effect of expert numbers}
We conduct ablation studies on the number of experts in the ESFC module, with results shown in Table~\ref{tab_esfc1}. The experiments demonstrate that 3 experts achieve the optimal balance between performance and computational efficiency, yielding the highest AP of 33.1\% and AP$_{50}$ of 52.7\% with marginal parameter increase. Using fewer experts (2) results in insufficient feature diversity, leading to 0.5\% AP drop, while more experts (4-5) introduce redundancy without performance gains and increase computational overhead. This validates our design choice of 3 experts as the sweet spot for adaptive feature extraction, providing sufficient representational capacity while maintaining parameter efficiency.

\begin{table}[ht]
\centering
\small 
\setlength{\tabcolsep}{5pt} 
\caption{Experiments with ESFC at different stages of feature fusion, where \textbf{D}, \textbf{M}, and \textbf{S} denote the Deep, Middle, and Shallow stages, respectively.}
\label{tab_esfc2}
\begin{tabular}{c|ccc|cc}
\toprule
\textbf{Stage} & \textbf{AP$^{val}$} & \textbf{AP$^{val}_{50}$} & \textbf{AP$^{val}_s$} & \textbf{Params(M)} & \textbf{GFLOPs} \\
\midrule
\textbf{S}        & 31.3 & 50.6 & 23.2 & 27.4 & 222.5 \\
\textbf{M}        & 32.5 & 52.1 & \textbf{24.9} & 27.4 & 277.4 \\
\textbf{M$\&$D}  & 32.3 & 51.5 & 24.4 & 25.9  & 272.8 \\
\textbf{D}        & \textbf{33.1} & \textbf{52.7} & 24.8 & 27.3 & 291.5\\
\bottomrule
\end{tabular}

\end{table}

\paragraph{Effect of ESFC at different fusion stages}

As shown in Table~\ref{tab_esfc2}, applying ESFC at the Deep stage yields the best performance, achieving an AP of 33.1\% and AP$_{50}$ of 52.7\%, surpassing the Shallow stage by 1.8\% and 2.1\%, respectively. Although the Middle stage demonstrates the highest accuracy for small object detection (AP$_s$ = 24.9\%), the Deep stage still achieves comparable results (24.8\%) while delivering superior overall accuracy. The M\&D configuration provides a parameter-efficient compromise with only 25.9M parameters, albeit with a slightly reduced AP of 32.3\%. These indicate that ESFC is most effective when applied in deeper layers, since it can better exploit semantic feature representations.

\paragraph{Effect of FFR design}

\begin{table}[!t]
\centering
\small 
\setlength{\tabcolsep}{5pt} 
\caption{Comparison with vs without high-level semantic feature in FFR. FFR$_{\textbf{F}_5}$ denotes designing of FFR retaining \textbf{F}$_5$.}
\label{tab_ffr}
\begin{tabular}{c|ccc|cc}
\toprule
\textbf{Method} & \textbf{AP$^{val}$} & \textbf{AP$^{val}_{50}$} & \textbf{AP$^{val}_s$} & \textbf{Params(M)} & \textbf{GFLOPs} \\
\midrule
FFR$_{\textbf{F}_5}$  & 30.1 & 49.6 & 22.1 & 30.5 & 242.0 \\
FFR  & \textbf{31.3} & \textbf{50.6} & \textbf{23.2} & 27.7 & 239.6 \\
\bottomrule
\end{tabular}

\end{table}

As presented in Table~\ref{tab_ffr}, we conduct ablation studies on different FFR designs based on the baseline. The results demonstrate that FFR enhances both performance and efficiency by eliminating  high-level semantic feature \textbf{F}$_5$ to mitigate redundancy in semantic information.

\section{Conclusion}
In this work, we propose EFSI-DETR, a novel detection framework that integrates efficient semantic feature enhancement with dynamic frequency-spatial guidance, tailored for challenging aerial scenarios. Specifically, we design two core modules: 1) DyFusNet, which jointly leverages frequency and spatial cues for robust multi-scale feature learning. 2) ESFC, which facilitates efficient semantic feature concentration. Furthermore, FFR strategy is adopted to preserve fine-grained spatial information. Extensive experiments demonstrate that our EFSI-DETR obtains superior performance while maintaining high-speed inference, particularly excelling in the detection of small and densely distributed objects.

\section*{Acknowledgments}
This work was supported by the NSFC Regional Innovation and Development Joint Fund under Grant U25A20537, the Shenzhen Science and Technology Program under Grant JCYJ20240813111301003, JCYJ20230807090103008, and Guangdong Basic and Applied Basic Research Foundation under Grant 2024A1515010456.

\bibliographystyle{IEEEtran}
\bibliography{name} 
\vspace{-1.0cm}
\begin{IEEEbiography}[{\includegraphics[width=1.0\textwidth]{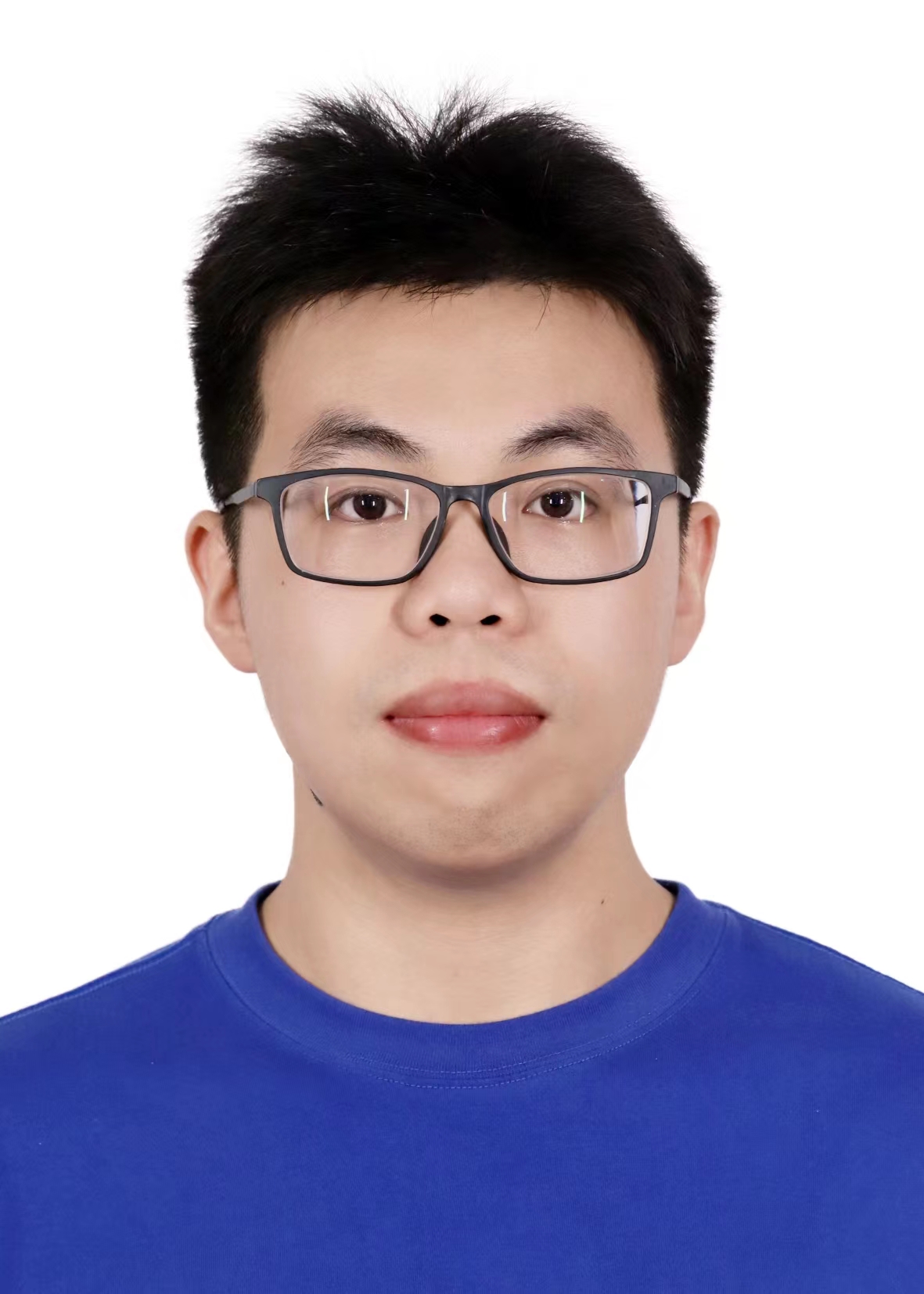}}]
{\textbf{Yu Xia}} received the B.S. degree from Northeastern University, Shenyang, China, in 2021, and the M.S. degree from Jiangnan University, Wuxi, China, in 2024. He is currently pursuing the Ph.D. degree in the State Key Laboratory of Information Engineering in Surveying, Mapping and Remote Sensing at Wuhan University. His research interests include computer vision, image processing, object detection, and human action recognition.
\end{IEEEbiography}

\vspace{-1.0cm}

\begin{IEEEbiography}[{\includegraphics[width=1.0\textwidth]{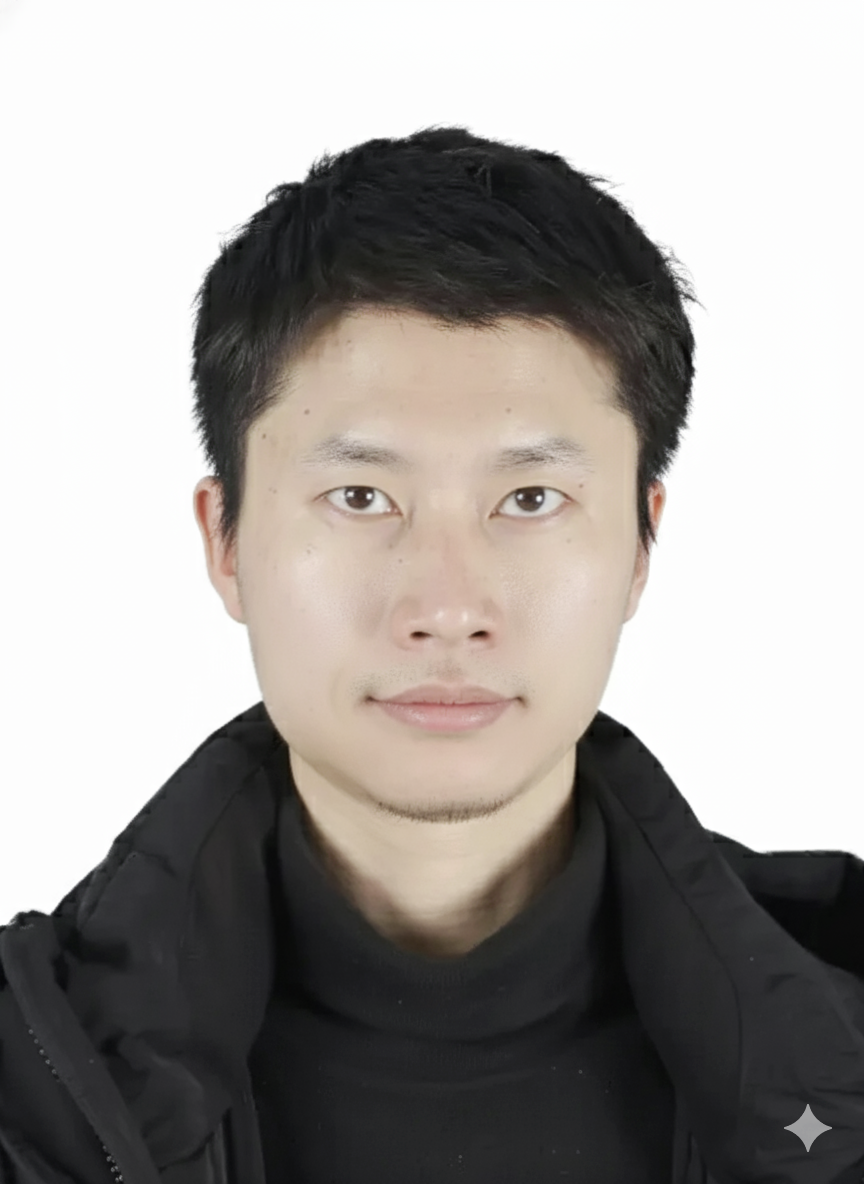}}]
{\textbf{Chang Liu}} is currently working toward the PhD degree with the School of Computer Science, Wuhan University, Wuhan, China. His research interests include computer vision and pattern recognition.
\end{IEEEbiography}

\vspace{-1.0cm}

\begin{IEEEbiography}[{\includegraphics[width=1.0\textwidth]{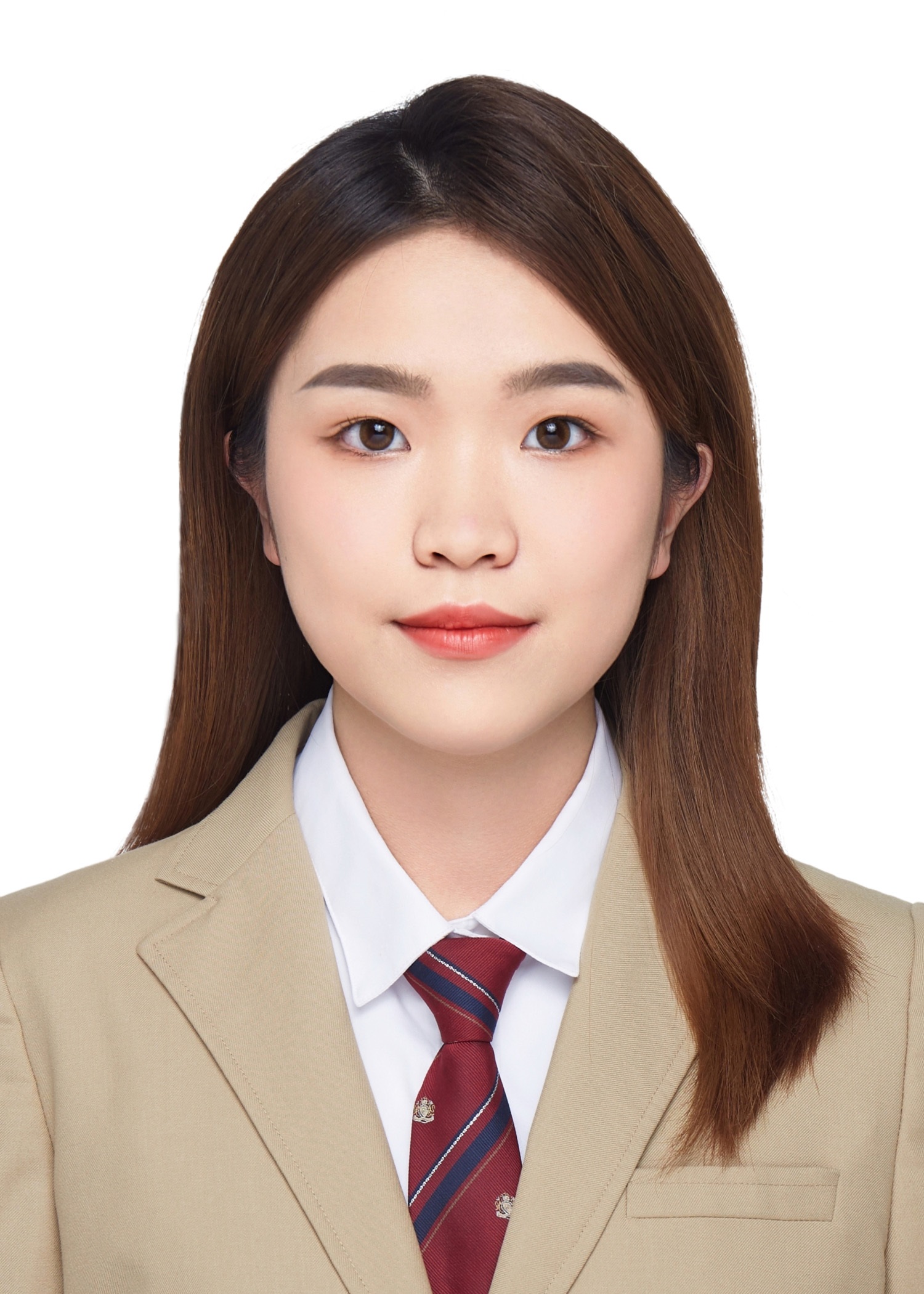}}]
{\textbf{Tianqi Xiang}} received the Bachelor’s degree in engineering from Xidian University, China, in 2023, and is currently pursuing a Master’s degree in the State Key Laboratory of Information Engineering in Surveying, Mapping and Remote Sensing at Wuhan University. Her research interests include computer vision, image processing, video analytics, object detection, and human action recognition.
\end{IEEEbiography}

\vspace{-1.0cm}

\begin{IEEEbiography}[{\includegraphics[width=1.0\textwidth]{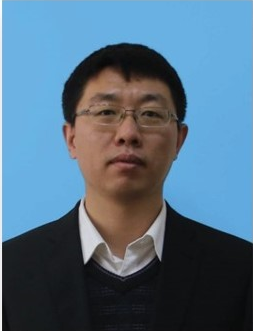}}]
{\textbf{Yang Cong}} (Senior Member, IEEE) received the B.Sc. degree from Northeast University in 2004 and the Ph.D. degree from the State Key Laboratory of Robotics, Chinese Academy of Sciences, in 2009. From 2009 to 2011, he was a Research Fellow with the National University of Singapore (NUS) and Nanyang Technological University (NTU). He was a Visiting Scholar with the University of Rochester. He was a Professor with Shenyang Institute of Automation, Chinese Academy of Sciences, until 2023. He is currently a Full Professor with South China University of Technology. He has authored more than 150 technical articles. His current research interests include robot, computer vision, machine learning, multimedia, medical imaging, and data mining. He has served on the editorial board of the several journal articles.
\end{IEEEbiography}

\vspace{-1.0cm}

\begin{IEEEbiography}[{\includegraphics[width=1.0\textwidth]{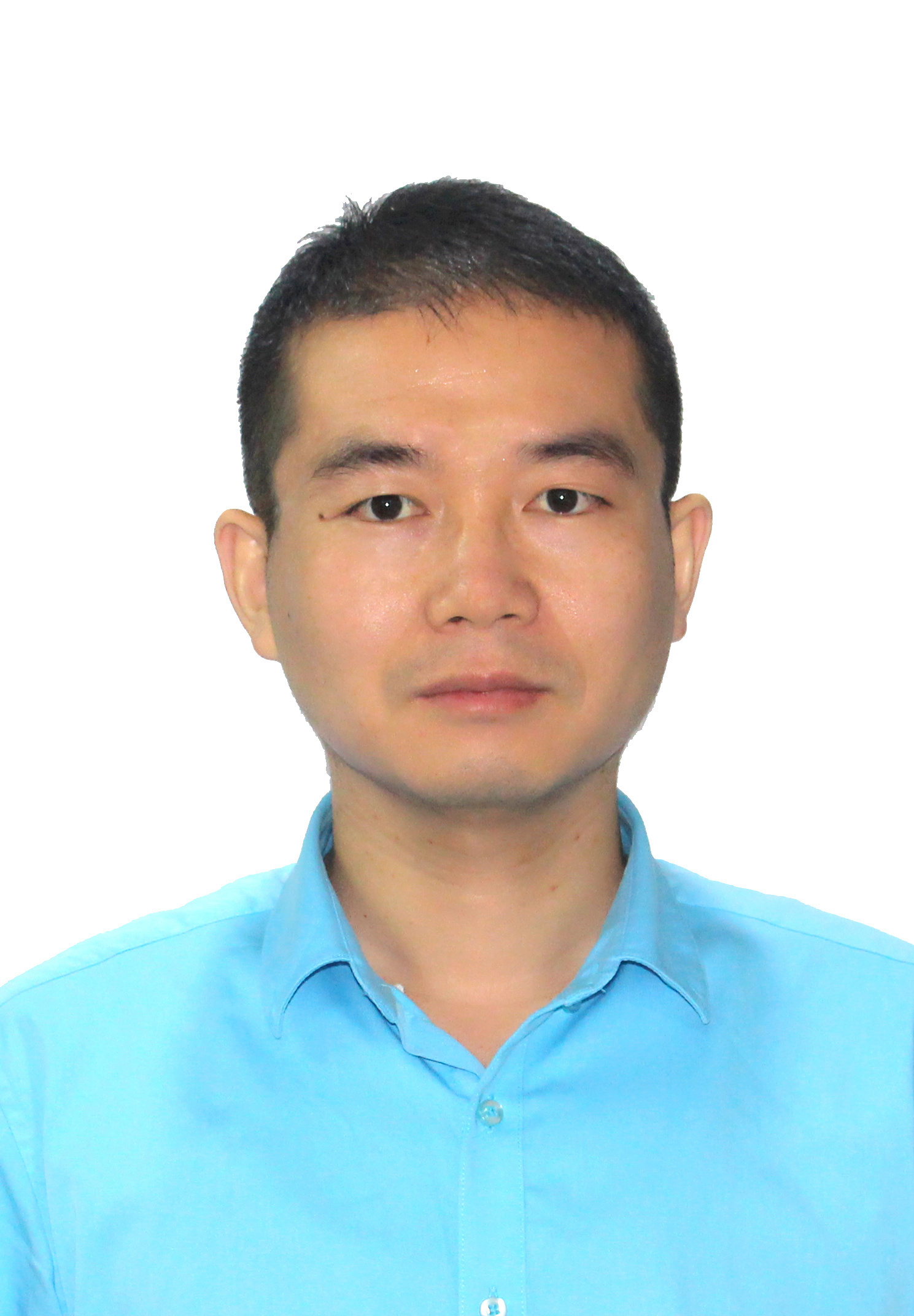}}]
{\textbf{Zhigang Tu}} (Senior Member, IEEE) received the Ph.D. degree from Wuhan University, China, 2013, and the Ph.D. degree from Utrecht University, Netherlands, 2015. From 2015 to 2016, he was a Post-doctor with Arizona State University, USA. From 2016 to 2018, he was a Research Fellow with Nanyang Technological University, Singapore.

He is currently a Professor with Wuhan University, and has co-/authored more than 100 papers in international SCI-indexed journals and conferences. His research interests include computer vision, image processing and video analytics, with focusing on motion estimation/retargeting, human behavior (action, pose, gesture) recognition and generation. He is the Area Chair of AAAI2023/2024/2025, Associate Editor of the SCI-indexed journals of CAAI Transactions on Intelligence Technology, Visual Computer, Journal of Visual Communications and Image Representation. He received the Best Student Paper Award at the 4th Asian Conference on Artificial Intelligence Technology, and one of the three Best Reviewers Award for IEEE Transactions on Circuits and Systems for Video Technology (IEEE T-CSVT) in 2022.
\end{IEEEbiography}
\end{document}